# Cost-Performance Optimization for Processing
# Low-Resource Language Tasks Using Commercial LLMs

**Arijit Nag**
IIT Kharagpur
arijitnag@iitkgp.ac.in

**Animesh Mukherjee**
IIT Kharagpur
animeshm@cse.iitkgp.ac.in

**Niloy Ganguly**
IIT Kharagpur
niloy@cse.iitkgp.ac.in

**Soumen Chakrabarti**
IIT Bombay
soumen@cse.iitb.ac.in

## Abstract

Large Language Models (LLMs) exhibit impressive zero/few-shot inference and generation quality for high-resource languages (HRLs). A few of them have been trained on low-resource languages (LRLs) and give decent performance. Owing to the prohibitive costs of training LLMs, they are usually used as a network service, with the client charged by the count of input and output tokens. The number of tokens strongly depends on the script and language, as well as the LLM's subword vocabulary. We show that LRLs are at a pricing disadvantage, because the well-known LLMs produce more tokens for LRLs than HRLs. This is because most currently popular LLMs are optimized for HRL vocabularies. Our objective is to level the playing field: reduce the cost of processing LRLs in contemporary LLMs while ensuring that predictive and generative qualities are not compromised. As means to reduce the number of tokens processed by the LLM, we consider code-mixing, translation, and transliteration of LRLs to HRLs. We perform an extensive study using the IndicXTREME classification and six generative tasks dataset, covering 15 Indic and 3 other languages, while using GPT-4 (one of the costliest LLM services released so far[1]) as a commercial LLM. We observe and analyze interesting patterns involving token count, cost, and quality across a multitude of languages and tasks. We show that choosing the best policy to interact with the LLM can reduce cost by ∼90% while giving better or comparable performance, compared to communicating with the LLM in the original LRL.

## 1 Introduction

Large Language Models (LLMs) like GPT-4 (OpenAI et al., 2023), ChatGPT, Llama-2 (Touvron et al., 2023), PaLM (Chowdhery et al., 2022), *inter alia*, are greatly contributing to the advancement of NLP with their exceptional zero/few-shot inference and generation abilities.

LLMs can also reduce dependence on expensive human-generated gold data for finetuning models for various downstream tasks, particularly for LRLs where gold data is scarce. However, our experiment suggests that this benefit is offset by one problem. Commercial LLM services like GPT-4 charge by the number of tokens exchanged with the client. The typical message from the client to the LLM consists of a *task description* or *instruction*, followed by zero or more *few-shot* "in-context examples", and the payload *instance* to be solved. The output tokens from the LLM express the solution to the payload instance.

The number of tokens exchanged between the client and LLM service depends on the (subword) vocabulary of the LLM, and the language(s) that the client uses. If the client's language is a LRL, chances are, the LLM will heavily segment the LRL tokens into subwords, because LRL subwords have minority status in most popular LLMs today (Hong et al., 2021). Consequently, use of the LLM will be expensive compared to HRL clients.

Despite giant leaps in reasoning and instruction-following capabilities, GPT-4 effectively discriminates against LRL clients in the same manner, and for the same reason, as its predecessor multilingual models mBERT and XLM-R (Conneau et al., 2020). This results in LRL clients being disadvantaged in terms of pricing, if not also quality. Our goal is to reduce this inequity.

Our choice of GPT-4 is driven by the fact that it is one of the costliest (yet most popular) commercial blackbox LLM-as-a-service. The cost of using the GPT-4 8K context model API is $0.03 per 1,000 input tokens and $0.06 per 1,000 output tokens. For the 32K context model, the cost is as high as $0.06 per 1,000 input tokens and $0.12 per 1,000 output tokens. Thus, a task of summarizing the Wikipedia (∼6 billion tokens) to half its size

---

[1] http://tinyurl.com/llm-costing

would cost $720,000 and $360,000 with GPT-4 having context length of 32K and 8K respectively.

Our initial study shows that GPT-4 generates diverse numbers of tokens for translations of a source sentence in different Indian languages. We also get a profile of default LRL task performances in these languages, if directly communicated to the LLM. The tool at our disposal is to preprocess the LRL messages sent to LLM service, in the face of a black-box service that we cannot influence. Specifically, we try to decrease the API cost by reducing the number of tokens exchanged between the client and LLM, while trying to maintain task quality. This can be done in various ways, as shown in Figure 1 and detailed later.

To compare these approaches, we define (in equation 1) a metric called the *RelaTive Performance to Cost Ratio* (RTPCR), which specifies the task performance we can get, given the token-driven cost we pay using various preprocessing methods $M$, relative to using the original LRL text. We desist from directly comparing the cost of LLM service access against operating costs of an in-house LLM, because the latter depends on too many hard-to-model factors like power and cooling costs (and comparing against the LLM service).

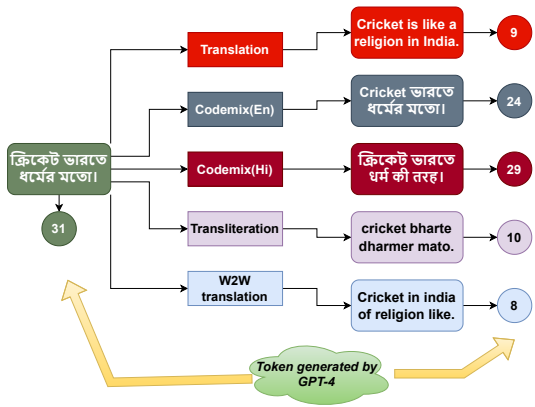

Figure 1: Different numbers of tokens generated by GPT-4 before and after preprocessing a LRL sentence using various techniques.

Explicit instance translation using LLM before inference will accrue extra cost, compared to free-of-charge translation using an open-source machine translation tool. An alternative is *implicit translation*, in which we instruct the LLM to translate the payload instance to English in the background before solving it, thus saving explicit translation cost. (We do not ask the LLM for the translated version.) We observe that the improvement using implicit translation trails behind ex-plicit translation. Therefore, using a dedicated LRL-to-HRL translator outside the LLM service can save cost *and* improve LRL task quality. A problem may arise if the LRL does not have a high-quality, open-sourced translator available and falling back on the LLM for translation will increase costs and decrease RTPCR. We next strike out in a *third* direction: *can mixing LRL and HRL words help*? To reduce cost, we can use a LRL-HRL or LRL-LRL dictionary, because such translations are often context-independent. Lastly, as a natural companion to selective wordmixing, we experiment with a *transliteration* technique, in which we transliterate the whole LRL message into Latin script. After all, the giant corpora used to train LLMs are more likely to contain LRL words heavily transliterated into Latin script, wordmixed with native LRL and HRL words, than 'pure' LRLs.

We summarize our contributions. (1) We identify a key cost-performance disadvantage faced by LLM clients that wish to solve LRL problems. (2) We demonstrate that this disadvantage is present in GPT-4 with respect to 15 Indian languages for diverse classification and generation tasks, using a new measure we introduce, called RTPCR. (3) We experiment with implicit and explicit client message translation and compare their RTPCR scores, to show that a free translator can reduce cost *and* improve performance. (4) We experiment with wordmixing and transliteration and show that both help reduce cost, but transliteration may result in modest performance impacts. (5) Apart from aggregative performance, we also present many case studies with interesting findings: The number of token counts varies greatly across Indian languages; translation to English may both increase and decrease the number of English words, depending on the source language; and LLM performance is rather sensitive to syntax.

## 2   Related Work

Recently LLMs like ChatGPT, GPT-4 (OpenAI et al., 2023), Llama-2 (Touvron et al., 2023), PaLM (Chowdhery et al., 2022), BLOOM (Scao et al., 2023), etc, have shown incredible zero-shot and few-shot capabilities across language and tasks. Although their performance in high-resource languages (HRLs) is impressive, studies (Hendy et al., 2023; Jiao et al., 2023; Bang et al., 2023) find that the same level of performance is not seen for low-resource languages (LRLs). Various benchmark datasets like XTREME (Hu et al., 2020), XTREME-R (Ruder et al., 2021), and XGLUE (Liang et al., 2020) have been designed to gauge the cross-lingual capabilities of a multilingual LLM. Benchmarks are also available in LRLs, e.g., IndicXTREME (Doddapaneni et al., 2023) for Indian languages. Benchmarks are also available in other language families like African (Adelani et al., 2022) and Indonesian (Wilie et al., 2020). In this work, we work with IndicXTREME. MEGA (Ahuja et al., 2023) witnesses improved task performance when translating the test instance to English. Having said that, they experiment from the performance perspective using the Bing translator(not an open-sourced tool). Unlike them, our focus is not only on performance but also on the cost. As inference using GPT-4 is costly for LRLs, we experiment with different techniques to reduce cost without hurting the performance. Similarly, a recent study (Huang et al., 2023) finds that a prompt written in English performs better than one written in other low-resource languages. Frugal-GPT (Chen et al., 2023) tries to address the cost aspect of GPT-4 on a different level, like selecting the fewer but more effective in-context examples for few-shot inference, caching previous queries and responses for future use, or LLM-cascading where cheap LLMs are tried first to check if the response is reasonable and only go for expensive ones if previous responses are not satisfactory. They have experimented with English, where the words are not fragmented much. LRLs present a steeper challenge, where first we need to reduce the cost by controlling over-fragmentation of LRL words. FrugelGPT can be applied thereafter to further reduce the cost. As per our knowledge, we are first to study the effect on performance when prepossessing the LRL input instance in a variety of different ways and relating it to the LLM API cost.

## 3 Methodology

In this work, we primarily preprocess the original LRL input instances using the techniques below to reduce the number of tokens generated.

### 3.1 Native

Here, we pass the query instance as it is in the native language script. We compare this scheme with all those stated below.

### 3.2 Translation

Here, we translate the LRL input instances before passing through the GPT-4. To translate, we consider with three possibilities.

**Using open-source MT:** We rely on off-the-shelf machine translation tools to translate LRL to English. We use IndicTrans2 (Gala et al., 2023) for this purpose.

**Using GPT-4 explicitly:** We first use GPT-4 as a translator to translate the LRL sentences to English; we then pass those translated English sentences to GPT-4 for inference.

**Using GPT-4 implicitly:** Here, our prompt (see Appendix Figure 6) instructs GPT-4 to translate LRL instances to English "within its premises" if it faces difficulty in understanding the LRL instances. The hope is that some cost will be saved because we do not ask for the English translation. The main problem with implicit translation is that we cannot control which LRL instances it chooses to translate, or the quality or effects of translation. Between the two explicit translation approaches, from the point of view of cost, obviously, the open-source MT tool is better, but it might be possible that for some LRLs, such MT may not be readily available. In such cases, explicit translation using GPT-4 can help, but it will come with the extra cost burden of API calling to do the translation beforehand.

### 3.3 Wordmix

In this approach, if a LRL word is excessively fragmented wrt the LLM subword vocabulary (see Algorithm 1 in Appendix B for details), then we will replace that word with its corresponding English translation. We call the new input **Wordmix(En)** as it is now a 'synthetic' text that contains a mix of LRL and English (or a more advantaged LRL, say Hindi — **Wordmix(Hi)**) words thus giving it the name. Note that, in the wordmix approach, we do not need any translation tool; an LRL-HRL or LRL-LRL dictionary suffices. The extreme case is where all LRL words are individually translated to English — **W2W**. For results of Wordmix(Hi) and W2W see Appendix A.

### 3.4 Transliteration

We transliterate[2] the LRL script into English (Latin script) in this approach. As we will now process the original LRL input instance in Latin script, the hypothesis is that fewer tokens shall be generated (as shown in Figure 1) than the original LRL script. Consequently, the cost should decrease. Another motivation behind using transliteration is that the training corpus of GPT-4 (and other popular LLMs, although mostly undisclosed) is likely to include social media or marketplace data, which includes a great deal of LRL text in Latin script.

---
[2] https://github.com/libindic/indic-trans

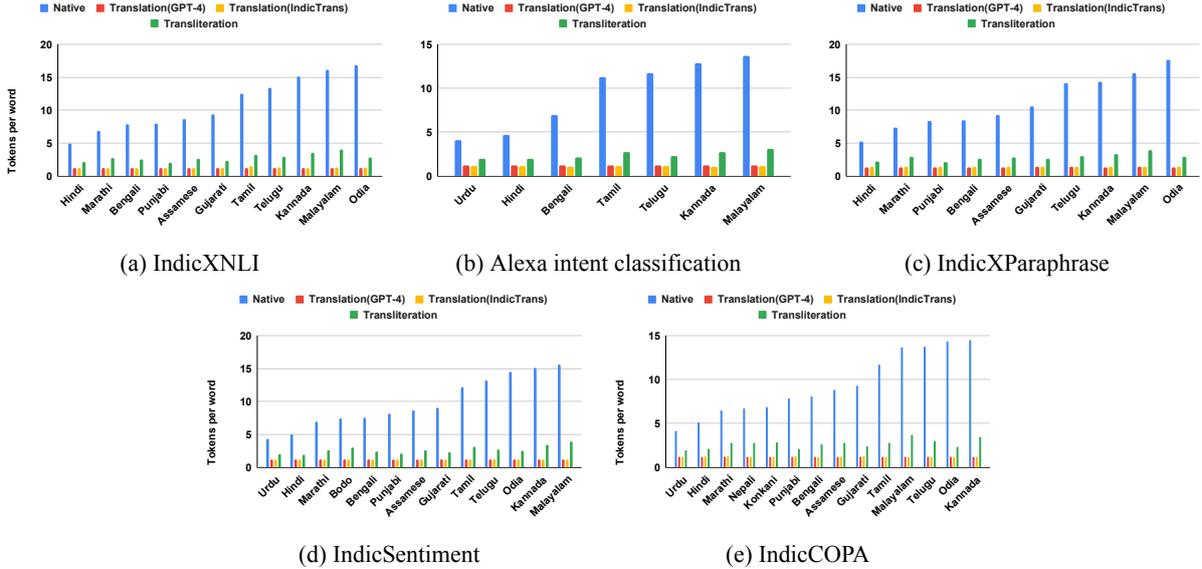

(a) IndicXNLI    (b) Alexa intent classification    (c) IndicXParaphrase

(d) IndicSentiment    (e) IndicCOPA

Figure 2: Average token generated per word by GPT-4. Here, we show the average number of tokens generated per word by GPT-4 when a sentence is processed in its own script (Native), translated via GPT-4 (labeled: Translation(GPT-4)), translated via open-sourced machine translation tool (labeled: Translation(IndicTrans)), and transliterated in English (labeled: Transliteration) for different Indian languages across various tasks.

### 3.5 RTPCR

To rank these techniques relative to using native script wrt both cost and performance, we devised a metric called RTPCR (RelaTive Performance to Cost Ratio):

$$\text{RTPCR}(M) = \frac{\frac{\text{Perf}(M)}{\text{Perf}(\textit{Native})}}{\frac{\text{Cost}(M)}{\text{Cost}(\textit{Native})}} \qquad (1)$$

Here $\text{Perf}(M)$ and $\text{Perf}(\textit{Native})$ denote the task-specific performance of the method $M$ and native LRL script, respectively. Similarly, $\text{Cost}(M)$ and $\text{Cost}(\textit{Native})$ represent the cost incurred by the method $M$ and native LRL script, respectively. RTPCR will be high when the performance of $M$ is better than native, and the cost is lower for $M$ than native. Conversely, a low RTPCR value indicates that $M$ lags in terms of performance but accrues more cost than using native text. Table 5 shows RTPCR values of different techniques.

## 4 Experiments and results

### 4.1 Datasets and evaluation metric

We experiment with all the classification tasks available in the IndicXTREME (Doddapaneni et al., 2023) benchmark dataset and six generative datasets which include summarization (XLSUM (Hasan et al., 2021)), Question-Answering(Chaii (Addison Howard, 2021), TyDi QA (Clark et al.)), Machine Translation(Samanantar (Ramesh et al., 2022)) and mathematical reasoning tasks (MGSM (Shi et al., 2022), Conic10K (Wu et al., 2023) and ArMath (Alghamdi et al., 2022)) total covering 15 Indian languages along with Thai, Swahili and Chinese. (details in Table 8). We report the accuracy score as a performance metric for classification tasks, ROUGE for summarization, BLEU for machine translation and Exact Match for question-answering and mathematical reasoning tasks. To make the GPT-4's output consistent and deterministic as much as possible, we set the seed and temperature parameters constant throughout the experiments. The Appendix discusses all the details related to different prompt designs (Figure 6) and GPT-4 API hyperparameters (Table 9).

### 4.2 Latin script reduces token generation

In Figure 2, we show the average number of tokens generated per word if a sentence passes through the

Table 1:

| Dataset | Techniques | As | Bn | Bd | Gu | Hi | Kn | Ko | Ml | Mr | Ne | Od | Pn | Ta | Te | Ur | Avg |
|---|---|---|---|---|---|---|---|---|---|---|---|---|---|---|---|---|---|
| **IndicSentiment** | Transliteration | -11.42 | -14.53 | 110.90 | 0.08 | -3.57 | 8.00 | - | -13.20 | -2.81 | - | 0.94 | -1.22 | -39.71 | -10.18 | -2.66 | -7.11 |
| | Translation(GPT-4) | 2.56 | 2.09 | 710.44 | 4.92 | **1.50** | 9.67 | - | 6.70 | 4.88 | - | 14.12 | <u>2.64</u> | 7.42 | 3.90 | **2.07** | 7.93 |
| | Translation(IndicTrans) | <u>5.91</u> | **2.93** | 1906.26 | **6.86** | 0.12 | **12.84** | - | <u>7.98</u> | 5.98 | - | <u>19.98</u> | 2.53 | **7.86** | **13.10** | 0.66 | **14.72** |
| | Translation(implicit) | 1.62 | 2.81 | 1362.65 | 1.75 | 0.67 | 8.79 | - | 6.70 | 0.09 | - | 12.69 | 3.88 | 7.85 | 2.09 | 1.26 | 9.59 |
| | Wordmix(En) | -2.24 | -1.35 | 488.63 | -6.45 | 0.70 | -0.38 | - | 1.11 | -3.37 | - | -8.59 | -3.31 | 3.04 | -7.27 | -0.92 | -0.33 |
| **IndicXParaphrase** | Transliteration | -9.36 | -17.17 | - | -20.07 | <u>0.56</u> | -20.72 | - | -22.06 | -6.90 | - | -13.69 | -2.74 | - | -21.11 | - | -13.50 |
| | Translation(GPT-4) | -0.27 | -6.55 | - | -2.38 | -1.12 | -4.27 | - | -4.66 | -1.22 | - | -2.82 | <u>**1.35**</u> | - | -1.27 | - | -2.49 |
| | Translation(IndicTrans) | <u>**5.91**</u> | -6.74 | - | -1.53 | -7.78 | -5.60 | - | -8.12 | -4.61 | - | -4.72 | 0.71 | - | -0.34 | - | -3.63 |
| | Translation(implicit) | 3.67 | <u>1.33</u> | - | -2.50 | 0.19 | <u>0.91</u> | - | **0.72** | **1.40** | - | <u>3.50</u> | 0.71 | - | -0.34 | - | **0.96** |
| | Wordmix(En) | 0.92 | -0.69 | - | -12.66 | -0.50 | -5.61 | - | -1.21 | -1.83 | - | -1.68 | -2.03 | - | <u>0.86</u> | - | -2.37 |
| **IndicCOPA** | Transliteration | -22.04 | -28.13 | - | -18.77 | -5.99 | -17.54 | -3.63 | -23.07 | -16.75 | -22.22 | -23.16 | -23.33 | -23.73 | -25.93 | -15.41 | -19.28 |
| | Translation(GPT-4) | -54.93 | -39.28 | - | -41.23 | -32.62 | -44.37 | -60.96 | -51.92 | -36.11 | -39.17 | -56.31 | -44.81 | -31.28 | -40.07 | -36.46 | -43.33 |
| | Translation(IndicTrans) | **13.54** | -7.80 | - | -8.40 | -23.85 | 7.03 | 4.14 | -10.24 | 1.24 | -22.89 | -15.63 | -22.48 | <u>21.65</u> | 0.59 | -23.86 | -7.54 |
| | Translation(implicit) | 8.69 | **10.94** | - | <u>1.50</u> | <u>12.11</u> | **22.82** | - | **20.32** | **9.00** | **22.27** | - | <u>**4.13**</u> | 15.00 | **11.11** | **5.23** | **10.32** |
| | Wordmix(En) | -3.40 | 9.38 | - | -7.25 | 5.10 | 8.78 | 10.91 | 7.39 | 0.67 | -10.27 | -14.13 | -13.62 | 5.07 | 10.28 | -0.87 | 0.08 |
| **IndicXNLI** | Transliteration | -22.28 | -20.18 | - | -8.81 | -7.45 | -18.66 | - | -19.71 | -10.02 | - | -16.90 | -11.69 | -30.57 | -14.97 | - | -16.35 |
| | Translation(GPT-4) | -14.72 | -2.26 | - | 0.30 | 0.79 | -1.56 | - | -0.10 | -6.02 | - | -9.10 | -5.24 | -40.07 | -7.66 | - | -4.96 |
| | Translation(IndicTrans) | **7.81** | **6.61** | - | <u>16.52</u> | <u>4.26</u> | <u>9.22</u> | - | <u>10.56</u> | **7.74** | - | **15.19** | **7.94** | <u>9.22</u> | <u>13.47</u> | - | **9.72** |
| | Translation(implicit) | 3.35 | 3.41 | - | 6.15 | 2.39 | 3.80 | - | 3.99 | 2.23 | - | 4.77 | 3.61 | 3.52 | 2.25 | - | 3.57 |
| | Wordmix(En) | -8.15 | -6.86 | - | -7.94 | -2.09 | -13.45 | - | -6.21 | -3.37 | - | -13.99 | -6.70 | -7.33 | -10.88 | - | -7.79 |

Table 1: Classification tasks: the change in accuracy (in terms of %) wrt the native script. The best results are in **boldface** and <u>underlined</u>. The last column shows the average performance across languages.

| Dataset | Techniques | Bn | Gu | Hi | Mr | Ne | Pa | Ta | Te | Ur | Avg |
|---|---|---|---|---|---|---|---|---|---|---|---|
| **Summarization** | Transliteration | -22.50 | -11.93 | -8.90 | -14.47 | -17.28 | -35.52 | -42.98 | -32.73 | -17.97 | -22.70 |
| | Translation(GPT-4) | -33.71 | -33.42 | -25.38 | -28.34 | -45.13 | -28.21 | -51.35 | -53.85 | -21.15 | -35.62 |
| | Translation(IndicTrans) | -36.99 | -35.87 | -23.57 | -32.59 | -46.50 | -31.41 | -51.63 | -50.32 | -18.91 | -36.42 |
| | Translation(implicit) | -0.84 | -4.23 | -2.38 | **12.04** | <u>10.96</u> | <u>7.31</u> | -0.65 | -1.90 | <u>5.46</u> | **2.86** |
| | Wordmix(En) | 0.00 | -23.94 | -5.84 | -2.13 | -10.73 | -12.75 | -10.14 | -24.03 | -9.98 | -10.99 |

| Dataset | Techniques | As | Bn | Gu | Kn | Ml | Mr | Od | Pa | Ta | Te | Avg |
|---|---|---|---|---|---|---|---|---|---|---|---|---|
| **MT** | Transliteration | -13.55 | -11.49 | -7.40 | -16.02 | -13.45 | -17.31 | -6.38 | -3.32 | -3.72 | -11.86 | -10.45 |
| | Translation(GPT-4) | 4.21 | -15.86 | **3.67** | -2.05 | -24.46 | -4.41 | -22.92 | -23.11 | -9.30 | 2.33 | -9.19 |
| | Translation(IndicTrans) | <u>**11.05**</u> | <u>10.32</u> | 2.07 | 5.03 | -2.85 | 0.22 | -13.85 | -19.64 | -2.94 | <u>2.84</u> | -0.78 |
| | Translation(implicit) | 1.84 | -11.68 | 0.20 | **6.37** | <u>5.98</u> | 1.76 | <u>7.06</u> | 1.79 | <u>2.06</u> | | **1.42** |
| | Wordmix(En) | -7.63 | -1.51 | -0.20 | 3.49 | -4.35 | -0.88 | <u>3.44</u> | 0.71 | -0.29 | -6.96 | -1.42 |

| Dataset | Techniques | Bn | Hi | Ta | Te | Avg |
|---|---|---|---|---|---|---|
| **QA** | Transliteration | -5.32 | -6.60 | -6.26 | 8.40 | -2.45 |
| | Translation(GPT-4) | -15.56 | -10.34 | -21.81 | -14.57 | -15.57 |
| | Translation(IndicTrans) | -7.41 | -5.08 | <u>4.12</u> | -3.76 | -3.03 |
| | Translation(implicit) | <u>4.68</u> | -1.54 | 2.32 | **11.40** | <u>4.22</u> |
| | Wordmix(En) | -10.81 | -15.47 | -6.53 | -8.42 | -10.31 |

| Dataset | Techniques | Ar | Bn | Te | Th | Sw | Zh | Avg |
|---|---|---|---|---|---|---|---|---|
| **Reasoning** | Transliteration | -21.04 | -45.87 | -54.79 | -86.00 | - | -23.53 | -46.25 |
| | Translation(GPT-4) | -1.86 | <u>14.86</u> | 23.65 | **2.33** | <u>15.45</u> | -10.15 | <u>7.38</u> |
| | Translation(IndicTrans) | -5.00 | 0.93 | **52.28** | -30.62 | -4.77 | -2.80 | 1.67 |
| | Translation(implicit) | <u>0.74</u> | 0.00 | 5.34 | -12.32 | 4.84 | -1.49 | -0.48 |
| | Wordmix(En) | -0.72 | -12.77 | -0.90 | -23.97 | -9.13 | **9.06** | -6.41 |

Table 2: Generation tasks: the change in BLEU(MT)/ROUGE(Summarization)/Exact match(QA,Mathematical reasoning) (in terms of %) wrt the native script. The best results are in **boldface** and <u>underlined</u>. The last column shows the average performance across languages.

| Dataset | Techniques | As | Bn | Bd | Gu | Hi | Kn | Ko | Ml | Mr | Ne | Od | Pn | Ta | Te | Ur | Avg |
|---|---|---|---|---|---|---|---|---|---|---|---|---|---|---|---|---|---|
| **IndicSentiment** | Transliteration | -69.78 | -68.34 | -59.68 | -75.28 | -61.24 | -77.11 | - | -75.12 | -61.36 | - | -83.27 | -73.84 | -74.14 | -79.63 | -53.34 | -72.44 |
| | Translation(GPT-4) | 33.28 | 36.40 | 43.80 | 27.34 | 42.76 | 21.89 | - | 23.06 | 42.37 | - | 17.42 | 26.64 | 27.31 | 33.16 | 44.67 | 29.00 |
| | Translation(IndicTrans) | **-83.69** | **-82.17** | **-79.81** | **-86.72** | **-79.33** | **-89.20** | - | **-88.73** | **-79.49** | - | **-91.62** | **-86.96** | **-86.84** | **-88.57** | **-78.84** | **-85.96** |
| | Wordmix(En) | -17.47 | -17.07 | - | -8.39 | -35.60 | -8.17 | -28.5 | - | -17.44 | - | -36.15 | -24.28 | -9.35 | -26.94 | -12.13 | -21.12 |
| **IndicXParaphrase** | Transliteration | -69.20 | -68.75 | - | -75.56 | -58.63 | -76.72 | - | -74.51 | -59.42 | - | -83.67 | -75.14 | - | -78.52 | - | -73.92 |
| | Translation(GPT-4) | 36.85 | 42.67 | - | 31.77 | 52.05 | 27.62 | - | 28.18 | 47.96 | - | 19.49 | 31.28 | - | 28.31 | - | 32.12 |
| | Translation(IndicTrans) | **-82.25** | **-79.29** | - | **-85.08** | **-75.48** | **-86.71** | - | **-86.40** | **-77.39** | - | **-90.58** | **-84.78** | - | **-86.38** | - | **-84.61** |
| | Wordmix(En) | -15.55 | -15.21 | - | -35.09 | -11.30 | -28.91 | - | -16.12 | -14.51 | - | -32.07 | -27.51 | - | -35.77 | - | -24.62 |
| **IndicCOPA** | Transliteration | -67.95 | -66.70 | - | -73.44 | -57.45 | -75.10 | -58.03 | -73.02 | -57.42 | -57.31 | -73.04 | -75.08 | -77.31 | -53.00 | - | -70.41 |
| | Translation(GPT-4) | 37.00 | 38.28 | - | 32.23 | 47.97 | 25.22 | 48.23 | 26.39 | 49.49 | 47.28 | 19.09 | 30.43 | 28.6 | 25.69 | 53.23 | 33.21 |
| | Translation(IndicTrans) | **-83.41** | **-82.30** | - | **-84.51** | **-76.59** | **-87.69** | **-76.74** | **-87.00** | **-76.22** | **-76.37** | **-90.46** | **-85.13** | **-87.09** | **-87.46** | **-73.85** | **-84.05** |
| | Wordmix(En) | -15.10 | -14.43 | - | -28.69 | -10.71 | -20.42 | -8.83 | -12.72 | -11.01 | -0.93 | -26.59 | -31.41 | -12.68 | -14.24 | -9.85 | -17.06 |
| **IndicXNLI** | Transliteration | -69.15 | -68.02 | - | -74.93 | -57.55 | -76.55 | - | -74.92 | -60.10 | - | -80.43 | -74.24 | -73.61 | -78.34 | - | -73.67 |
| | Translation(GPT-4) | 33.58 | 36.53 | - | 27.63 | 43.15 | 22.84 | - | 23.6 | 41.92 | - | 17.03 | 26.91 | 27.17 | 24.46 | - | 27.57 |
| | Translation(IndicTrans) | **-84.14** | **-82.70** | - | **-86.69** | **-78.46** | **-88.81** | - | **-88.54** | **-79.54** | - | **-91.68** | **-86.82** | **-86.49** | **-87.83** | - | **-86.56** |
| | Wordmix(En) | -11.36 | -8.76 | - | -29.12 | -8.79 | -32.13 | - | -15.80 | -10.62 | - | -30.56 | -26.47 | -12.64 | -33.57 | - | -21.89 |

Table 3: Classification tasks: the change in token generation per instance (in terms of %) wrt the native script is shown. The best results are in **boldface** and <u>underlined</u>. The last column shows the average cost across languages.

GPT-4 in its own script, is translated to English using open-sourced MT and GPT-4, and is transliterated to English script. The general observation is that token generation is greatly reduced when processing the instance in Latin script (translation or transliteration). The reduction can be anywhere between 2× and 7×, depending on the languages. Another interesting observation is that the reduction is higher for Dravidian languages like Malayalam, Tamil, Telugu, Kannada, etc., while the reduction is comparatively less in North-Indian languages like Hindi and Bengali. This might imply that GPT -4 is better trained in North-Indian languages than South-Indian ones. Another thing to notice here is that although transliteration does not necessarily produce a valid English word, it still helps to reduce the number of token generation.

**Table 4 — Summarization**

| Dataset | Techniques | Bn | Gu | Hi | Mr | Ne | Pa | Ta | Te | Ur | Avg |
|---|---|---|---|---|---|---|---|---|---|---|---|
| Summarization | Transliteration | -67.81 | -73.39 | -57.75 | -59.72 | -59.23 | -75.08 | -70.37 | -73.55 | -51.90 | -65.42 |
| | Translation(GPT-4) | 12.15 | 10.75 | 19.65 | 13.28 | 18.92 | 11.91 | 12.98 | 10.36 | 24.91 | 14.99 |
| | **Translation(IndicTrans)** | **-82.84** | **-84.94** | **-79.10** | **-78.43** | **-79.55** | **-86.02** | **-84.64** | **-83.94** | **-75.72** | **-81.69** |
| | Wordmix(En) | -11.42 | -25.23 | -10.94 | -11.31 | -9.48 | -28.73 | -7.42 | -13.64 | -10.79 | -14.33 |

**Table 4 — MT**

| Dataset | Techniques | As | Bn | Gu | Kn | Ml | Mr | Od | Pa | Ta | Te |
|---|---|---|---|---|---|---|---|---|---|---|---|
| MT | Transliteration | -56.08 | -55.46 | -59.11 | -58.37 | -57.84 | -57.04 | -56.91 | -56.77 | -56.42 | -57.21 / -57.12 |
| | Translation(GPT-4) | 43.27 | 45.09 | 47.74 | 49.09 | 48.54 | 45.81 | 49.43 | 47.84 | 45.53 | 48.86 / 47.12 |
| | **Translation(IndicTrans)** | **-78.60** | **-76.92** | **-76.55** | **-77.42** | **-76.76** | **-77.06** | **-76.71** | **-76.65** | **-77.83** | **-75.92 / -77.04** |
| | Wordmix(En) | 0.00 | 0.00 | 0.00 | -2.65 | -1.27 | -11.03 | 0.00 | -0.01 | 0.00 | 0.67 / -1.43 |

**Table 4 — QA**

| Dataset | Techniques | Bn | Hi | Ta | Te | Avg |
|---|---|---|---|---|---|---|
| QA | Transliteration | -68.09 | -56.70 | -70.94 | -75.97 | -67.93 |
| | Translation(GPT-4) | 39.16 | 45.40 | 11.61 | 33.30 | 32.37 |
| | **Translation(IndicTrans)** | **-78.81** | **-74.79** | **-84.48** | **-81.79** | **-79.97** |
| | Wordmix(En) | -15.28 | -10.40 | -33.29 | -36.32 | -23.82 |

**Table 4 — Reasoning**

| Dataset | Techniques | Ar | Bn | Te | Th | Sw | Zh | Avg |
|---|---|---|---|---|---|---|---|---|
| Reasoning | Transliteration | -35.87 | -66.26 | -78.81 | -54.95 | - | 35.83 | -40.01 |
| | Translation(GPT-4) | 94.10 | 42.02 | 25.35 | 59.93 | 115.78 | 185.40 | 87.10 |
| | **Translation(IndicTrans)** | **-56.22** | **-79.56** | **-87.08** | **-75.25** | **-42.71** | **-5.86** | **-57.78** |
| | Wordmix(En) | -20.46 | -19.06 | -45.50 | -37.10 | -4.08 | -2.32 | -21.42 |

Table 4: Generation tasks: the change in token generation per instance (in terms of %) wrt the native script is shown. The best results are in **boldface** and underlined. The last column shows the average cost across languages.

**Table 5**

| Dataset | Techniques | As | Bn | Bd | Gu | Hi | Kn | Ko | Ml | Mr | Ne | Od | Pn | Ta | Te | Ur | Avg |
|---|---|---|---|---|---|---|---|---|---|---|---|---|---|---|---|---|---|
| IndicSentiment | Transliteration | 2.93 | 2.70 | 5.23 | 4.05 | 2.49 | 4.72 | - | 3.49 | 2.52 | - | 6.03 | 3.78 | 2.33 | 4.41 | 2.09 | 3.37 |
| | Translation(GPT-4) | 0.77 | 0.75 | 5.64 | 0.82 | 0.71 | 0.90 | - | 0.87 | 0.74 | - | 0.97 | 0.81 | 0.84 | 0.84 | 0.71 | 0.84 |
| | **Translation(IndicTrans)** | **6.49** | **5.77** | **99.38** | **8.05** | **4.84** | **10.45** | - | **9.58** | **5.17** | - | **14.32** | **7.86** | **8.20** | **9.90** | **4.76** | **8.17** |
| | Translation(implicit) | 1.02 | 1.03 | 14.63 | 1.02 | 1.01 | 1.09 | - | 1.07 | 1.00 | - | 1.13 | 1.04 | 1.08 | 1.02 | 1.01 | 1.10 |
| | Wordmix(En) | 1.18 | 1.11 | 6.43 | 1.45 | 1.10 | 1.39 | - | 1.22 | 1.06 | - | 1.43 | 1.28 | 1.14 | 1.27 | 1.13 | 1.26 |
| IndicXParaphrase | Transliteration | 2.94 | 2.65 | - | 3.27 | 2.43 | 3.40 | - | 3.06 | 2.29 | - | 5.28 | 3.91 | - | 3.67 | - | 3.32 |
| | Translation(GPT-4) | 0.73 | 0.65 | - | 0.74 | 0.65 | 0.75 | - | 0.74 | 0.67 | - | 0.81 | 0.77 | - | 0.77 | - | 0.74 |
| | **Translation(IndicTrans)** | **5.97** | **4.50** | - | **6.60** | **3.76** | **7.10** | - | **6.76** | **4.22** | - | **10.11** | **6.62** | - | **7.32** | - | **6.26** |
| | Translation(implicit) | 1.04 | 1.01 | - | 0.97 | 1.00 | 1.01 | - | 1.01 | 1.01 | - | 1.04 | 1.01 | - | 1.00 | - | 1.01 |
| | Wordmix(En) | 1.19 | 1.12 | - | 1.35 | 1.12 | 1.33 | - | 1.18 | 1.15 | - | 1.45 | 1.35 | - | 1.57 | - | 1.30 |
| IndicCOPA | Transliteration | 2.43 | 2.16 | - | 3.06 | 2.21 | 3.31 | 2.30 | 2.85 | 1.96 | 1.82 | 4.55 | 2.84 | 3.06 | 3.34 | 1.80 | 2.73 |
| | Translation(GPT-4) | 0.33 | 0.44 | - | 0.44 | 0.46 | 0.44 | 0.26 | 0.38 | 0.43 | 0.41 | 0.37 | 0.42 | 0.53 | 0.48 | 0.41 | 0.43 |
| | **Translation(IndicTrans)** | **6.84** | **5.21** | - | **5.91** | **3.25** | **8.70** | **4.48** | **6.90** | **4.26** | **3.26** | **9.03** | **5.21** | **9.42** | **8.02** | **2.91** | **5.79** |
| | Translation(implicit) | 1.09 | 1.11 | - | 1.01 | 1.12 | 1.23 | 1.20 | 1.09 | 1.22 | 1.11 | 0.97 | 1.04 | 1.15 | 1.11 | 1.05 | 1.10 |
| | Wordmix(En) | 1.14 | 1.28 | - | 1.30 | 1.18 | 1.37 | 1.22 | 1.23 | 1.13 | 0.91 | 1.17 | 1.26 | 1.20 | 1.29 | 1.10 | 1.21 |
| IndicXNLI | Transliteration | 2.52 | 2.50 | - | 3.64 | 2.18 | 3.47 | - | 3.20 | 2.25 | - | 5.02 | 3.43 | 2.63 | 3.93 | - | 3.18 |
| | Translation(GPT-4) | 0.64 | 0.72 | - | 0.79 | 0.70 | 0.80 | - | 0.81 | 0.66 | - | 0.78 | 0.75 | 0.7 | 0.74 | - | 0.74 |
| | **Translation(IndicTrans)** | **6.80** | **6.16** | - | **8.75** | **4.84** | **9.76** | - | **9.65** | **5.27** | - | **13.84** | **8.19** | **8.08** | **9.32** | - | **8.16** |
| | Translation(implicit) | 1.03 | 1.03 | - | 1.06 | 1.02 | 1.04 | - | 1.04 | 1.02 | - | 1.05 | 1.04 | 1.04 | 1.02 | - | 1.04 |
| | Wordmix(En) | 1.04 | 1.02 | - | 1.30 | 1.07 | 1.28 | - | 1.11 | 1.08 | - | 1.24 | 1.27 | 1.06 | 1.34 | - | 1.18 |

Table 5: RTPCR values for all techniques across all classification datasets and languages. The best results are in **boldface** and underlined. The last column shows the average RTPCR across languages.

### 4.3 Translation performs better

Table 1 and 2 shows the change in performance (%) compared to the native script for classification and generation tasks, respectively (complete result in Appendix Table 10 & 11). For classification, it is clear that, on the whole, translation improves the GPT-4 performance, be it translation by GPT-4, open-source MT like IndicTrans or implicit translation by GPT-4. However, translation by IndicTrans has the upper hand most of the time. From this observation, it is clear that GPT-4 better understands text in English than LRLs. Next, Wordmix(En) also shows promise by giving a comparable performance to native script, given that our wordmix technique is a bit crude, and further improvement can be possible by improving the word replacement heuristics. Wordmix(Hi) also performs comparable to Wordmix(En), but the latter is more fruitful when considering the cost reduction (details in Appendix A). On the other hand, in transliteration, performance drops, and it is expected as we are not translating the word to English but changing the script. For generation tasks, except the reasoning tasks, direct translation does not seem to be beneficial from the performance point of view, but here, the implicit translation improves the performance of Summarization, MT and QA tasks.

| Dataset | Techniques | Bn | Gu | Hi | Mr | Ne | Pa | Ta | Te | Ur | | Avg |
|---|---|---|---|---|---|---|---|---|---|---|---|---|
| Summarization | Transliteration | 2.41 | 3.31 | 2.16 | 2.12 | 2.03 | 2.59 | 1.92 | 2.54 | 1.71 | | 2.31 |
| | Wordmix(En) | 1.13 | 1.02 | 1.06 | 1.10 | 0.99 | 1.22 | 0.97 | 0.88 | 1.02 | | 1.00 |
| | Translation(IndicTrans) | **3.67** | **4.26** | **3.66** | **3.12** | **2.62** | **4.91** | **3.15** | **3.09** | **3.34** | | **<u>3.54</u>** |
| | Translation(GPT-4) | 0.59 | 0.60 | 0.62 | 0.63 | 0.46 | 0.64 | 0.43 | 0.42 | 0.63 | | 0.56 |
| | Translation(implicit) | 0.99 | 0.97 | 0.98 | 1.12 | 1.11 | 1.07 | 1.00 | 0.99 | 1.04 | | 1.03 |
| | | **As** | **Bn** | **Gu** | **Kn** | **Ml** | **Od** | **Pa** | **Ta** | **Te** | | |
| MT | Transliteration | 1.97 | 1.99 | 2.26 | 2.02 | 2.05 | 1.92 | 2.17 | 2.24 | 2.21 | 2.06 | 2.09 |
| | Wordmix(En) | 0.92 | 0.98 | 1.00 | 1.06 | 0.97 | 1.11 | 1.03 | 1.01 | 1.00 | 0.92 | 1.00 |
| | Translation(IndicTrans) | **5.19** | **4.78** | **4.35** | **4.65** | **4.18** | 4.37 | **3.7** | 3.44 | **4.38** | **4.27** | **<u>4.33</u>** |
| | Translation(GPT-4) | 0.73 | 0.58 | 0.70 | 0.66 | 0.51 | 0.66 | 0.52 | 0.52 | 0.62 | 0.69 | 0.62 |
| | Translation(implicit) | 1.02 | 0.98 | 1.00 | 1.06 | 1.06 | 1.07 | 1.02 | 1.02 | 1.01 | 0.90 | 1.01 |
| | | **Bn** | **Hi** | **Ta** | **Te** | | | | | | | |
| QA | Transliteration | 2.97 | 2.16 | 3.23 | 4.51 | | | | | | | 3.22 |
| | Wordmix(En) | 1.05 | 0.94 | 1.40 | 1.44 | | | | | | | 1.21 |
| | Translation(IndicTrans) | **4.37** | **3.76** | **6.71** | **5.29** | | | | | | | **<u>5.03</u>** |
| | Translation(GPT-4) | 0.61 | 0.62 | 0.70 | 0.64 | | | | | | | 0.64 |
| | Translation(implicit) | 1.05 | 0.98 | 1.02 | 1.11 | | | | | | | 1.04 |
| | | **Ar** | **Bn** | **Te** | **Th** | **Sw** | **Zh** | | | | | |
| Reasoning | Transliteration | 1.23 | 1.60 | 2.13 | 0.31 | - | 0.56 | | | | | 1.17 |
| | Wordmix(En) | 1.25 | 1.08 | 1.82 | 1.21 | 0.95 | 1.12 | | | | | 1.24 |
| | Translation(IndicTrans) | **2.17** | **4.94** | **11.79** | **2.80** | **1.66** | **1.03** | | | | | **<u>4.07</u>** |
| | Translation(GPT-4) | 0.51 | 0.81 | 0.99 | 0.64 | 0.54 | 0.31 | | | | | 0.63 |
| | Translation(implicit) | 1.01 | 0.99 | 1.07 | 0.87 | 1.03 | 0.98 | | | | | 0.99 |

Table 6: RTPCR values for all techniques across all generation datasets and languages. The best results are in **boldface** and underlined. The last column shows the average RTPCR across languages.

### 4.4 Open-source MT reduces cost

In Table 3 and 4, we compare the cost incurred by different techniques for GPT-4 inference. As the cost is proportional to the tokens generated from the inputs and outputs, we use the average token generated per query instance as a proxy for cost. We show the percentage change in token generation compared to the native script (While reporting the average token generation per instance for different techniques in Appendix Table 12 & 13). Here, the observation is clear: all the techniques reduce cost except the translation via GPT-4. It is expected since in this method, we first explicitly translate the instance using GPT-4 and then use the translated instance for classification/generation again using GPT-4. So here, we need to take the extra burden of translation costs using GPT-4. On the other hand, translation via IndicTrans (open-source MT) reduces the cost by as much as 90%. But in case of unavailability of such free MT, one can go for wordmix or transliteration, which also reduces the cost significantly. Having said that, transliteration can impact performance (see Table 1 & 2).

### 4.5 RTPCR helps find best value for money

Table 5 and 6 shows RTPCR for different methods compared to the Native script. Here, the translation via IndicTrans is the clear winner for obvious reasons. The translation in this case is free, and GPT-4 performance as we see in Table 1 improved when using English-translated query instances. The lowest RTPCR score is for translation using GPT-4 because although the performance improves after translation, the cost we pay for translation outweighs the performance gain. RTPCR for transliteration is also high — though the performance drops a bit, it reduces the cost significantly compared to using native script. RTPCRs for wordmix and implicit translation via GPT-4 are also better than native (RTPCR is 1), as wordmix performs comparably as native but reduces cost. On the other hand, the cost of implicit translation by GPT-4 is the same as the native script, but the performance gain is more for implicit translation.

### 4.6 Analysis of the results

In Table 7, we show examples covering different cases in our methodology. The first two examples are related to the translation method. As shown earlier, translation improves the inference performance; however, this solely depends on the quality of the translation. In the first example, the original sentence in Bengali expresses a negative sentiment sarcastically; GPT-4 finds it challenging to comprehend it in Bengali and, as a result, produces a wrong classification. However, when we translate it to English (translation quality is good), GPT-4 can classify it correctly. In the second example, although the translation looks acceptable on the surface, it misses the negative emotion latent in the Hindi sentence, resulting in a wrong prediction. The subsequent two cases are related to the word-mix technique. Here, the first one talks about the terrible (Bn: ভয়ঙ্কর) storytelling of a movie and also uses the phase "সেরা উদাহরণ" (great example); GPT-4 focuses on the later to make a wrong positive prediction. Nevertheless, when we replace ভয়ঙ্কর with 'terrible', GPT-4 produces the correct result. One problem with this approach is that we replace any word in the original sentence with its translation without looking at the context, and it can sometimes produce non-meaningful sentences. The following example shows such a case where wordmixing produces an incoherent Odia-English wordmix sentence, and GPT-4 fails to predict anything. The last example corresponds to the implicit translation using the GPT-4 while inference. Here, the original sentence in Bengali produces the wrong prediction, but when we advise GPT-4 to translate the Bn sentence to En "in the background" if needed, it produces the correct result. To clarify, we do not output the translation explicitly to reduce the output token generation; this is just to verify and illustrate the technique's effectiveness.

| Cases | Instances | Labels |
|---|---|---|
| **Translation helps** | **Original sentence:** আপনি আপনার পছন্দের বিষয়ে সমালোচনা করতে পারেন তবে যেহেতু এটি একটি বিনামূল্যের প্ল্যাটফর্ম তাই যে কেউ এমনকি একটি বানরও এটি ব্যবহার করতে পারে। বানরটি অনেকের চেয়ে বেশি ভালো কাজ করতে পারতো। <br> **Translated sentence:** You can critique on your favorite topics, but since this is a free platform, anyone, even a monkey, can use it. The monkey could do a better job than many. | **Gold label:** Negative <br> **Native script prediction:** Positive <br> **Translation prediction:** Negative |
| **Translation hurts** | **Original sentence:** इसकी न्यूनतम क्षमता 1.5 टन है, जो कि100 sq. ft. केएक छोटे से कमरे के लिए बहुत अधिक है, जो आम तौर पर एक मध्यम वर्ग के घर में किसी भी जगह का क्षेत्र है। मेरा सारा पैसा ले लो!!! <br> **Translated sentence:** Its minimum capacity is 1.5 tons, <u>which is quite a lot for a small room</u> of 100 sq. ft., which is typically the area of any place in a middle-class home. Take all my money!!! | **Gold label:** Negative <br> **Native script prediction:** Negative <br> **Translation prediction:** Positive |
| **Wordmix helps** | **Original sentence:** এই সিনেমাটি ভয়ঙ্কর স্টোরি টেলিং-এর একটি সেরা উদাহরণ, তবে <u>পুরোটাই নেতিবাচক অর্থে</u>। প্রতিটি দৃশ্যকে বোঝার জন্য আপনাকে সময় বের করতে হবে। <br> **Translated sentence:** এই সিনেমাটি <u>Terrible</u> স্টোরি টেলিং-এর একটি সেরা উদাহরণ, তবে <u>All of it নেতিবাচক অর্থে</u>। প্রতিটি the view বোঝার জন্য আপনাকে সময় বের করতে হবে। | **Gold label:** Negative <br> **Native script prediction:** Positive <br> **wordmix prediction:** Negative |
| **Wordmix hurts** | **Original sentence:** ଶଢ଼ା ପ୍ଲାଷ୍ଟିକରେରଠାରୀ। ୟୁନ୍ଟାଲ୍ଟିବାରେବେଢୁଢ଼ଧ୍ୟଥ୍ରୁଢ଼ ଏବଂ ଶେଷରେ ଅର୍ଥିଇ ଅପ୍ରଜନ। <br> **Translated sentence:** ଶଢ଼ା <u>in the plastic.</u> ୟୁନ୍ଟାଲି। ୟୁନ୍ଟ୍ While you're running ବେଢୁଢ଼ଧ୍ୟଥ୍ରୁଢ଼ ଏବଂ ଶେଷରେ ଅର୍ଥିଇ ଅପ୍ରଜନ। | **Gold label:** Negative <br> **Native script prediction:** Negative <br> **wordmix prediction:** Failed to predict |
| **Implicit translation** | **Original sentence:** এটি প্রতিটি লোমের টুকরাকে বিচ্ছিন্ন করে দেয়। এটি সমস্ত আলগা লোম টেনে বের করে এবং ঝরে পড়া থেকে রক্ষা করে। <br> **Implicit translation:** এটি প্রতিটি লোমের টুকরাকে বিচ্ছিন্ন করে দেয়। এটি সমস্ত আলগা লোম টেনে বের করে এবং ঝরে পড়া থেকে রক্ষা করে।<translation>It separates every piece of hair. It pulls out all the loose hair and prevents it from falling out.</translation> | **Gold label:** Positive <br> **Native script prediction:** Neutral <br> **Translation prediction:** Positive |

Table 7: Case analysis. The first two examples show the translation technique's good and bad sides, followed by the next two examples related to the Wordmix(En) method. The last example shows how implicit translation can help with inference. (Note: To reduce cost we do not explicitly output the translation of the sentence; this is just for illustration purposes.) Detailed discussion is in section 4.6. Important words related to the sentiment of sentences are <u>underlined</u>.

## 5 Conclusion

We study the cost of solving five classification and six generation task datasets across various Indian languages using GPT-4, the leading commercial LLM. While GPT-4 does well on these tasks, LRL words get highly fragmented, leading to high cost of API calls. We wish to retain the performance advantage of GPT-4 while reducing API cost. To that end, we try different pre-processing techniques involving translation, wordmixing, and transliteration. We find that translating the whole LRL sentence to English reduces the cost and improves performance (given a free LRL→HRL MT system). In the absence of such an MT system, implicit translation using GPT-4 or even an LRL-HRL dictionary can be used to replace highly fragmented words and reduce costs. In future, we wish to explore wordmix more deeply while using other HRLs related to the source LRL.

## 6 Limitations

Despite high costs, GPT-4 is widely regarded as providing better quality than open-source counterparts like BLOOM (Scao et al., 2023) and GPT-neo. GPT-4 has been compared and shown to be superior to BARD in multiple application scenarios including text processing and understanding[3]. For this reason, and to contain costs, we only experiment with one commercial LLM, GPT-4. Although it will increase the experiment cost significantly, bringing other paid LLMs into the experimental setup will make our observations more robust. Also, our wordmix technique, where we replace an LRL word with its translation, without taking sentence context into account, is crude — it sometimes impacts sentence structure. Replacing a word or phrase while considering sentence context will be more appropriate. In this work, we only focus on Indian languages. Although we cover as many as 15 languages, an immediate extension can be to check the hypothesis for other LRLs worldwide.

## References

divy thakkar Julia Elliott Partha Talukdar Phil Culliton Addison Howard, Deepak Nathani. 2021. chaii - hindi and tamil question answering.

David Ifeoluwa Adelani et al. 2022. MasakhaNER 2.0: Africa-centric transfer learning for named entity recognition.

Kabir Ahuja, Harshita Diddee, Rishav Hada, Millicent Ochieng, Krithika Ramesh, Prachi Jain, Akshay Nambi, Tanuja Ganu, Sameer Segal, Maxamed Axmed, Kalika Bali, and Sunayana Sitaram. 2023. Mega: Multilingual evaluation of generative ai.

Reem Alghamdi, Zhenwen Liang, and Xiangliang Zhang. 2022. ArMATH: a dataset for solving Arabic math word problems. In *Proceedings of the Thirteenth Language Resources and Evaluation Conference*, pages 351–362, Marseille, France. European Language Resources Association.

Yejin Bang, Samuel Cahyawijaya, Nayeon Lee, Wenliang Dai, Dan Su, Bryan Wilie, Holy Lovenia, Ziwei Ji, Tiezheng Yu, Willy Chung, Quyet V. Do, Yan Xu, and Pascale Fung. 2023. A multitask, multilingual, multimodal evaluation of chatgpt on reasoning, hallucination, and interactivity.

Lingjiao Chen, Matei Zaharia, and James Zou. 2023. Frugalgpt: How to use large language models while reducing cost and improving performance.

Aakanksha Chowdhery et al. 2022. Palm: Scaling language modeling with pathways.

Jonathan H. Clark, Eunsol Choi, Michael Collins, Dan Garrette, Tom Kwiatkowski, Vitaly Nikolaev, and Jennimaria Palomaki. Tydi qa: A benchmark for information-seeking question answering in typologically diverse languages.

Alexis Conneau, Kartikay Khandelwal, Naman Goyal, Vishrav Chaudhary, Guillaume Wenzek, Francisco Guzmán, Edouard Grave, Myle Ott, Luke Zettlemoyer, and Veselin Stoyanov. 2020. Unsupervised cross-lingual representation learning at scale. In *Proceedings of the 58th Annual Meeting of the Association for Computational Linguistics*, pages 8440–8451, Online. Association for Computational Linguistics.

Sumanth Doddapaneni, Rahul Aralikatte, Gowtham Ramesh, Shreya Goyal, Mitesh M. Khapra, Anoop Kunchukuttan, and Pratyush Kumar. 2023. Towards leaving no indic language behind: Building monolingual corpora, benchmark and models for indic languages.

Jay Gala, Pranjal A Chitale, A K Raghavan, Varun Gumma, Sumanth Doddapaneni, Aswanth Kumar M, Janki Atul Nawale, Anupama Sujatha, Ratish Puduppully, Vivek Raghavan, Pratyush Kumar, Mitesh M Khapra, Raj Dabre, and Anoop Kunchukuttan. 2023. Indictrans2: Towards high-quality and accessible machine translation models for all 22 scheduled indian languages. *Transactions on Machine Learning Research*.

Tahmid Hasan, Abhik Bhattacharjee, Md. Saiful Islam, Kazi Mubasshir, Yuan-Fang Li, Yong-Bin Kang, M. Sohel Rahman, and Rifat Shahriyar. 2021. XL-sum: Large-scale multilingual abstractive summarization for 44 languages. In *Findings of the Association for Computational Linguistics: ACL-IJCNLP 2021*, pages 4693–4703, Online. Association for Computational Linguistics.

Amr Hendy, Mohamed Abdelrehim, Amr Sharaf, Vikas Raunak, Mohamed Gabr, Hitokazu Matsushita, Young Jin Kim, Mohamed Afify, and Hany Hassan Awadalla. 2023. How good are gpt models at machine translation? a comprehensive evaluation.

Jimin Hong, TaeHee Kim, Hyesu Lim, and Jaegul Choo. 2021. AVocaDo: Strategy for adapting vocabulary to downstream domain. In *Proceedings of the 2021 Conference on Empirical Methods in Natural Language Processing*, pages 4692–4700, Online and Punta Cana, Dominican Republic. Association for Computational Linguistics.

Junjie Hu, Sebastian Ruder, Aditya Siddhant, Graham Neubig, Orhan Firat, and Melvin Johnson. 2020. XTREME: A massively multilingual multitask benchmark for evaluating cross-lingual generalization.

---

[3]https://themeisle.com/blog/chatgpt-vs-google-bard/#gref

Haoyang Huang, Tianyi Tang, Dongdong Zhang, Xin Zhao, Ting Song, Yan Xia, and Furu Wei. 2023. Not all languages are created equal in LLMs: Improving multilingual capability by cross-lingual-thought prompting. In *Findings of the Association for Computational Linguistics: EMNLP 2023*, pages 12365–12394, Singapore. Association for Computational Linguistics.

Wenxiang Jiao, Wenxuan Wang, Jen tse Huang, Xing Wang, Shuming Shi, and Zhaopeng Tu. 2023. Is chatgpt a good translator? yes with gpt-4 as the engine.

Yaobo Liang et al. 2020. Xglue: A new benchmark dataset for cross-lingual pre-training, understanding and generation.

OpenAI et al. 2023. Gpt-4 technical report.

Gowtham Ramesh, Sumanth Doddapaneni, Aravinth Bheemaraj, Mayank Jobanputra, Raghavan AK, Ajitesh Sharma, Sujit Sahoo, Harshita Diddee, Mahalakshmi J, Divyanshu Kakwani, Navneet Kumar, Aswin Pradeep, Srihari Nagaraj, Kumar Deepak, Vivek Raghavan, Anoop Kunchukuttan, Pratyush Kumar, and Mitesh Shantadevi Khapra. 2022. Samanantar: The largest publicly available parallel corpora collection for 11 Indic languages. *Transactions of the Association for Computational Linguistics*, 10:145–162.

Sebastian Ruder, Noah Constant, Jan Botha, Aditya Siddhant, Orhan Firat, Jinlan Fu, Pengfei Liu, Junjie Hu, Dan Garrette, Graham Neubig, and Melvin Johnson. 2021. Xtreme-r: Towards more challenging and nuanced multilingual evaluation.

Teven Le Scao et al. 2023. Bloom: A 176b-parameter open-access multilingual language model.

Freda Shi, Mirac Suzgun, Markus Freitag, Xuezhi Wang, Suraj Srivats, Soroush Vosoughi, Hyung Won Chung, Yi Tay, Sebastian Ruder, Denny Zhou, Dipanjan Das, and Jason Wei. 2022. Language models are multilingual chain-of-thought reasoners.

Hugo Touvron et al. 2023. LLaMA 2: Open foundation and fine-tuned chat models.

Bryan Wilie et al. 2020. IndoNLU: Benchmark and resources for evaluating Indonesian natural language understanding. In *Proceedings of the 1st Conference of the Asia-Pacific Chapter of the Association for Computational Linguistics and the 10th International Joint Conference on Natural Language Processing*, pages 843–857, Suzhou, China. Association for Computational Linguistics.

Haoyi Wu, Wenyang Hui, Yezeng Chen, Weiqi Wu, Kewei Tu, and Yi Zhou. 2023. Conic10K: A challenging math problem understanding and reasoning dataset. In *Findings of the Association for Computational Linguistics: EMNLP 2023*, pages 6444–6458, Singapore. Association for Computational Linguistics.

# Cost-Performance Optimization for Processing Low-Resource Language Tasks Using Commercial LLMs
## (Appendix)

## A Supplementary results

| Dataset | Language covered | Example |
|---|---|---|
| IndicSentiment | Assamese, Bengali, Bodo, Gujarati, Hindi, Kannada, Malayalam, Marathi, Odia, Punjabi, Tamil, Telugu, Urdu | Sentence: The recently included feature of stories by defaultly visible for 24 hrs, makes me super happy. Social connection just at ease!<br>Label: Positive |
| IndicXParaphrase | Assamese, Bengali, Gujarati, Hindi, Kannada, Malayalam, Marathi, Odia, Punjabi, Telugu | Sentence 1: it is found in bangladesh, india, myanmar, and pakistan.<br>Sentence 2: it is not found in bangladesh, india, myanmar, and pakistan.<br>Label: No |
| IndicCOPA | Assamese, Bengali, Konkani, Gujarati, Hindi, Kannada, Malayalam, Marathi, Nepali, Odia, Punjabi, Tamil, Telugu, Urdu | Premise: The man broke his toe.<br>Choice 1: He got a hole in his sock.<br>Choice 2: He dropped a hammer on his foot.<br>Question: CAUSE<br>Label: Choice2 |
| IndicXNLI | Assamese, Bengali, Gujarati, Hindi, Kannada, Malayalam, Marathi, Odia, Punjabi, Tamil, Telugu | Premise: You don't have to stay here.<br>Hypothesis: You can leave.<br>Label: Entailment |
| Alexa Intent classification | Hindi, Bengali, Telugu, Urdu, Malayalam, Kannada, Tamil | Instruction: Please turn off the light.<br>Intent: lightoff. |
| XLSUM<br><br>(Summarization) | Bengali, Gujarati, Hindi, Marathi, Nepali, Panjabi, Tamil Telugu, Urdu | Paragraph: The Met Office has issued a yellow weather warning for wind covering Wales and England, starting from 21:00 GMT on Wednesday evening. Travel and power are both likely to be disrupted, with the warning to remain in place until 15:00 on Thursday. Gusts of 55mph (88kmh) are likely and could hit up to 70mph on coasts and hills, with heavy and blustery showers.<br>Summary: Winds could reach gale force in Wales with stormy weather set to hit the whole of the country this week. |
| Samanantar<br><br>(Machine Translation) | Hindi-> {Assamese, Bengali, Gujarati, Marathi, Malayalam, Kannada, Odia, Panjabi, Tamil, Telugu} | Source: वीडियो क्लिप शेयर किए<br>Target: Shared the video clip. |
| chaii, TyDi QA<br><br>(Question-Answering) | Hindi, Bengali, Telugu, Tamil | Context: Before the Delhi Durbar was built, the Gateway of India was built to commemorate the visit of King George V and Queen Mary to Mumbai in 1911. But they could only see a cardboard model of the structure, as construction began after 1915. [11] Sir George Sydenham Clarke, the Government of Bombay, laid the foundation stone for the architecture on 31 March 1911. The final design was approved on March 31, 1913. The gateway was paved with yellow basalt and concrete. [12] The foundation's work was completed in 1920 and the entire work in 1924. [13] Viceroy The Earl of Reading opened the gateway on 4 December 1924. [6].<br>Question: Who laid the foundation stone of the Gateway of India architecture built in the city of Mumbai in the western Indian state of Maharashtra?.<br>Answer: Sir George Sydenham Clarke |
| MGSM, Conic_math, Ar_math<br><br>(Mathematical reasoning) | Arabic, Bengali, Telugu, Thai, Swahili, Chinese | Question: Janet's ducks lay 16 eggs per day. She eats three for breakfast every morning and bakes muffins for her friends every day with four She sells the remainder at the farmers' market daily for $2 per fresh duck egg. How much in dollars does she make every day at the farmers market?.<br>Answer: 18. |

Table 8: Dataset details. Examples (in English) are for illustration; actual datasets are in Indian languages.

## Datasets

**Distinct words reduces after translation**  In Figure 3, we plot the change in distinct words in an LRL sentence before and after the translation using both the GPT-4 and open-source MT tools. Surprisingly, we observe that for most of the languages, the number of distinct words drops significantly after translation for both methods. The number of distinct words increased after translation only for languages – Hindi, Urdu, and Punjabi. One thing that needs to be noted here is that the reduction of distinct words is more for the South-Indian languages than other Indic languages. This can potentially lead to a loss of information, and as an effect, the model's performance after translation can suffer.

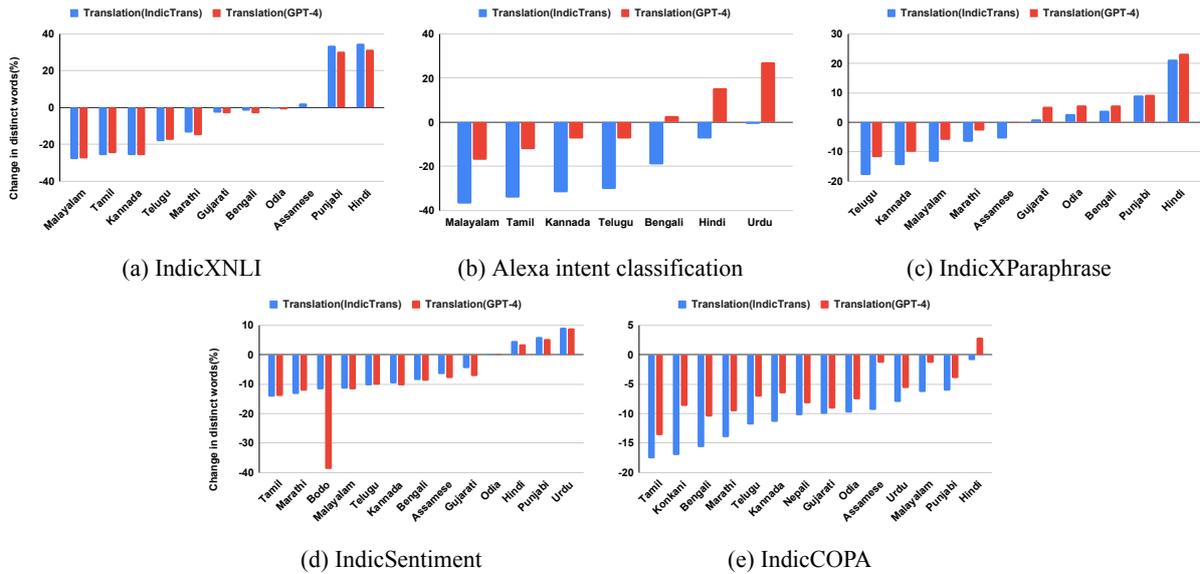

Figure 3: Change in distinct words after translation. Here, we compare change in distinct words after translation using (IndicTrans and GPT-4) with sentence in native scripts for different Indian languages across tasks.

**Wordmix(Hi)** In place of English it is also possible to use a more advantaged LRL. We did our experiments with Hindi as such a choice since among the Indian LRLs, Hindi words are comparatively less fragmented (as shown in Figure 2). We call this variant **Wordmix(Hi)**.

**Effect of wordmixing** In Figure 4, we check the effect of wordmixing in more detail. We compare two different wordmixing techniques involving English and Hindi as target languages to translate an LRL word. While the choice of English is understandable, we choose Hindi because it is fragmented less compared to other Indian languages (shown in Figure 2) and also with the assumption that a language from the same language family can cooperate better in a sentence. As per observation, although Wordmix(Hi) cannot reduce the *cost* as compared to Wordmix(En), Wordmix(Hi) often *performs* better than Wordmix(En). To some extent, this validates a hypothesis of effective transfer between LRL members of the same language family. Having said that, these observations demand a deeper exploration involving different target languages to get a deeper understanding.

**W2W (Dictionary)** Here, we take a LRL-English dictionary and replace each LRL word with its corresponding English translation. It is quite obvious that W2W will hamper the sentence syntax and semantics, but having said that, we use it as we are getting a free approximate translation of the LRL sentence and also to check if word order really matters in GPT-4 inference.

**Sentence structure matters** In Figure 5, we compare the performance of GPT-4 between techniques that import English words in the original native script sentence. These techniques change the structure or syntax of the sentence differently, unlike the full-sentence translation using MT, which keeps the sentence syntax intact. For instance, wordmix can change sentence structure as it heuristically replaces any word with its corresponding English translation, totally oblivious of the context. Similarly, the extreme case can be W2W translation, where we use a dictionary to replace each word in the original sentence with its corresponding English word. This can produce a completely incoherent sentence without proper syntax. Our experiment shows out of these three techniques, total translation using MT always does better, followed by Wordmix(En). W2W gives the worst performance. This establishes the fact that although English script is much easier for GPT-4 to understand, word structure matters.

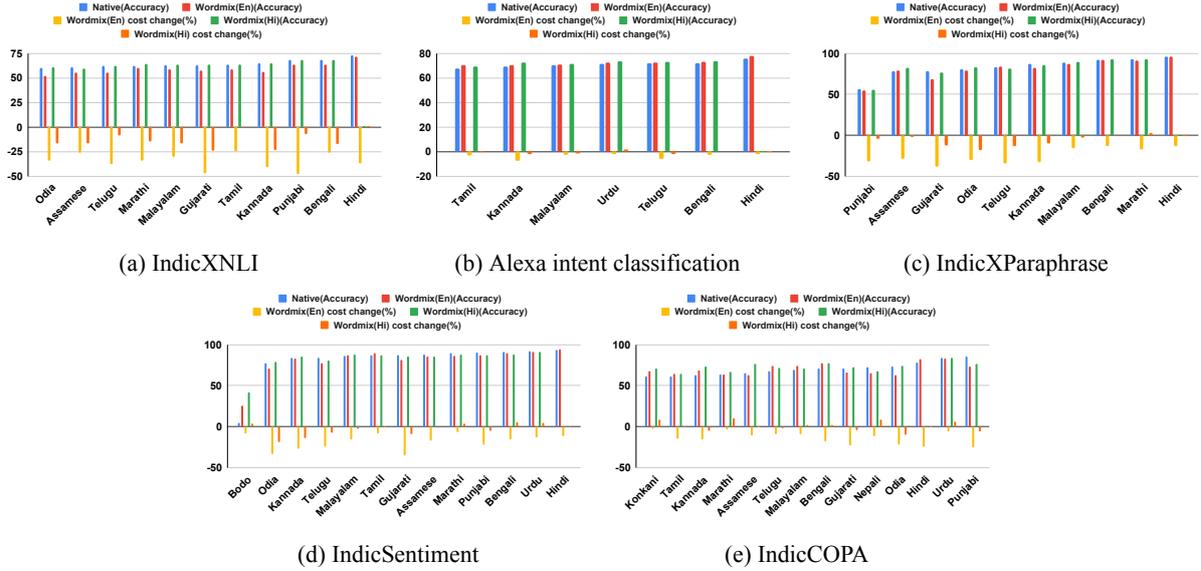

(a) IndicXNLI    (b) Alexa intent classification    (c) IndicXParaphrase

(d) IndicSentiment    (e) IndicCOPA

Figure 4: GPT-4 wordmixing performance vs cost reduction comparison while mixing native script with an HRL (English) and LRL (Hindi). Here, we plot the accuracy using native script, Wordmix(Native-En), and Wordmix(Native-Hi) sentence inference along with the change of API cost compared to native script inferencing for different Indian languages across tasks.

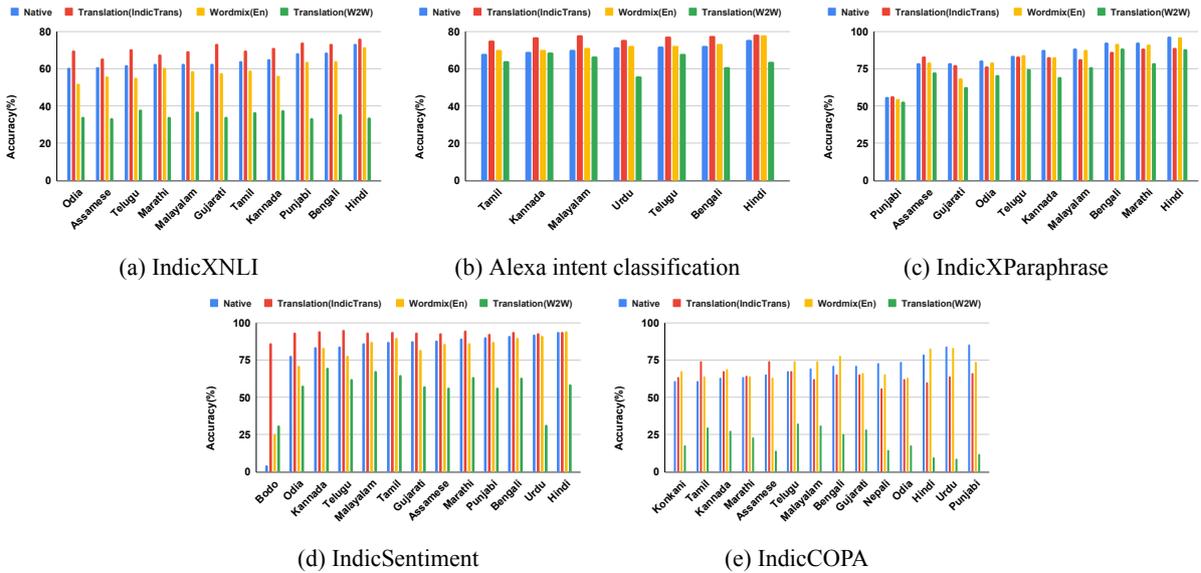

(a) IndicXNLI    (b) Alexa intent classification    (c) IndicXParaphrase

(d) IndicSentiment    (e) IndicCOPA

Figure 5: Impact on different ways of incorporating English words while GPT-4 inferencing. Here, we compare performance using native script, full sentence translation using MT (IndicTrans), partial native word replacement to English (Wordmix(En)), and full word-to-word translation (W2W) using LRL-HRL dictionary for different Indian languages across tasks.

## B   Word selection algorithm

## C   Experimental settings

| Hyperparameter | Value |
|---|---|
| LLM | GPT-4-0613 |
| GPT4 temperature | 0.5 (during translation), 0.0001(during classification), 0.1(during generation) |
| GPT4 max token | 256 |
| GPT4 Seed | 42 |

Table 9: Details of GPT-4 hyperparameters.

We use GPT-4-0613 model, which costs $0.03 / 1K tokens for input and $0.06 / 1K tokens for output. We run all the experiments in 16GB RAM CPU based system, without any GPU usage. To preserve cost,

---

**Algorithm 1** wordmix: Word selection.

---

    **inputs** LRL corpus $D$, LLM tokenizer $\mathcal{T}$, LRL-HRL dictionary $Dic_{LtH}$

    **outputs** wordmix LRL corpus $D_{CM}$

1:   $D_{CM} \leftarrow$ empty list

2:   $L_W \leftarrow$ empty list                                           ▷ *Word length list*

3:   $SR_W \leftarrow$ empty list                                 ▷ *Token to word length split ratio list*

4:   $\mathcal{W} \leftarrow$ distinct words from corpus $D$

5:   **for** $w \in \mathcal{W}$ **do**

6:       $L_W \leftarrow len(w)$

7:       $SR_W \leftarrow \frac{len(\mathcal{T}(w))}{len(w)}$

8:   **for** $s \in D$ **do**

9:       **for** $w \in s$ **do**

10:         **if** $len(w) \geq mean(L_W)$ **&** $\frac{len(\mathcal{T}(w))}{len(w)} \geq mean(SR_W)$ **then**

11:            Replace $w$ with $Dic_{LtH}(w)$ in $s$

12:       $D_{CM} \leftarrow s$

    **return** $D_{CM}$

---

we do all the experiments one time, and to make them reproducible we fix the seed value to 42 and set temperature close to zero for classification tasks of the GPT-4 API.

**Native Script**

**Prompt description:** Classify the sentiment(positive/negative) of the Hindi sentence.

**Sentence:** वीडियो कॉलिंग के लिए, वे आपको अपना मैसेंजर इन्स्टॉल करने के लिए कहेंगे। तो मूल रूप से एक सोशल मीडिया के लिए, आपको 2 अलग-अलग ऐप्स की आवश्यकता है, क्या यह बकवास नहीं है?

**Label:**

**Translated/W2W instance**

**Prompt description:** Classify the sentiment(positive/negative) of the sentence.

**Sentence:** For video calling, they will ask you to install your messenger. So basically for a social media, you need 2 different apps, isn't that rubbish?

**Label:**

**Transliterated instance**

**Prompt description:** Classify the sentiment(positive/negative) of the Hindi to English transliterated sentence.

**Sentence:** video colling ke liye, ve aapako apana massenger inastol karne ke liye kahenge. to mul rup se ek soshal media ke liye, aapako 2 alag-alag eps kii aavashyaktaa he, kya yah bakawaas nahin he?

**Label:**

**Codemix instance**

**Prompt description:** Classify the sentiment(positive/negative) of the Hindi-English codemixed sentence.

**Sentence:** Video calling के लिए, वे आपको अपना Messenger Install करने के लिए will say. तो मूल रूप से एक सोशल मीडिया के लिए, आपको 2 different ऐप्स की Necessity है, क्या यह बकवास नहीं है?

**Label:**

**Implicit translation instance**

**Prompt description:** Classify the sentiment(positive/negative) of the Sentence. For your better understanding you can translate the Hindi Sentence to English in background when needed.

**Sentence:** वीडियो कॉलिंग के लिए, वे आपको अपना मैसेंजर इन्स्टॉल करने के लिए कहेंगे। तो मूल रूप से एक सोशल मीडिया के लिए, आपको 2 अलग-अलग ऐप्स की आवश्यकता है, क्या यह बकवास नहीं है?

**Label:**

Figure 6: Prompts used for different techniques. Here, all these prompts are related to the IndicSentiment dataset.

| Dataset | Techniques | As | Bn | Bd | Gu | Hi | Kn | Ko | Ml | Mr | Ne | Od | Pn | Ta | Te | Ur | Avg |
|---|---|---|---|---|---|---|---|---|---|---|---|---|---|---|---|---|---|
| **IndicSentiment** | Native script | 88.04 | 91.39 | 4.31 | 87.56 | 93.78 | 83.73 | - | 86.54 | 89.42 | - | 77.99 | 90.43 | 87.30 | 84.10 | 92.35 | 81.30 |
| | Transliteration | 77.99 | 78.11 | 9.09 | 87.63 | 90.43 | 90.43 | - | 75.12 | 86.91 | - | 78.72 | 89.33 | 52.63 | 75.54 | 89.89 | 75.52 |
| | | -11.42 | -14.53 | 110.90 | 0.08 | -3.57 | 8.00 | - | -13.20 | -2.81 | - | 0.94 | -1.22 | -39.71 | -10.18 | -2.66 | -7.11 |
| | Translation(GPT-4) | 90.29 | 93.30 | 34.93 | 91.87 | **95.19** | 91.83 | - | 92.34 | 93.78 | - | 89.00 | **92.82** | 93.78 | 87.38 | **94.26** | 87.75 |
| | | 2.56 | 2.09 | 710.44 | 4.92 | **_1.50_** | 9.67 | - | 6.70 | 4.88 | - | 14.12 | **_2.64_** | 7.42 | 3.90 | **2.07** | 7.93 |
| | Translation(IndicTrans) | **93.24** | **94.07** | **86.47** | **93.57** | 93.89 | **94.48** | - | **93.45** | **94.77** | - | **93.57** | 92.72 | **94.16** | **95.12** | 92.96 | **93.27** |
| | | **_5.91_** | **2.93** | **1906.26** | **6.86** | 0.12 | **12.84** | - | **7.98** | **5.98** | - | **19.98** | 2.53 | **7.86** | **13.10** | 0.66 | **14.72** |
| | Translation(implicit) | 89.47 | 93.96 | 63.04 | 89.09 | 94.41 | 91.09 | - | 92.34 | 89.50 | - | 87.89 | 93.94 | 94.15 | 85.86 | 93.51 | 89.10 |
| | | 1.62 | 2.81 | 1362.65 | 1.75 | 0.67 | 8.79 | - | 6.70 | 0.09 | - | 12.69 | 3.88 | 7.85 | 2.09 | 1.26 | 9.59 |
| | Wordmix(En) | 86.07 | 90.16 | 25.37 | 81.91 | 94.44 | 83.41 | - | 87.50 | 86.41 | - | 71.29 | 87.44 | 89.95 | 77.99 | 91.50 | 81.03 |
| | | -2.24 | -1.35 | 488.63 | -6.45 | 0.70 | -0.38 | - | 1.11 | -3.37 | - | -8.59 | -3.31 | 3.04 | -7.27 | -0.92 | -0.33 |
| **IndicXParaphrase** | Native Script | 78.67 | 92.50 | - | **78.67** | 96.68 | 87.53 | - | 88.64 | 92.80 | - | 80.56 | 56.11 | - | 83.57 | - | 83.57 |
| | Transliteration | 71.31 | 76.62 | - | 62.88 | **97.22** | 69.39 | - | 69.09 | 86.40 | - | 69.53 | 54.57 | - | 65.93 | - | 72.29 |
| | | -9.36 | -17.17 | - | -20.07 | **_0.56_** | -20.72 | - | -22.06 | -6.90 | - | -13.69 | -2.74 | - | -21.11 | - | -13.50 |
| | Translation(GPT-4) | 78.46 | 86.44 | - | 76.80 | 95.60 | 83.79 | - | 84.51 | 91.67 | - | 78.29 | **56.87** | - | 82.51 | - | 81.49 |
| | | -0.27 | -6.55 | - | -2.38 | -1.12 | -4.27 | - | -4.66 | -1.22 | - | -2.82 | **1.35** | - | -1.27 | - | -2.49 |
| | Translation(IndicTrans) | **83.38** | 86.27 | - | 77.47 | 89.16 | 82.63 | - | 81.44 | 88.52 | - | 76.76 | 56.51 | - | 83.29 | - | 80.54 |
| | | **_5.99_** | -6.74 | - | -1.53 | -7.78 | -5.60 | - | -8.12 | -4.61 | - | -4.72 | 0.71 | - | -0.34 | - | -3.63 |
| | Translation(implicit) | 81.56 | **93.73** | - | 76.70 | 96.86 | **88.33** | - | **89.28** | **94.10** | - | **83.38** | 56.51 | - | 83.29 | - | **84.37** |
| | | 3.67 | **1.33** | - | -2.50 | 0.19 | **0.91** | - | **0.72** | **1.40** | - | **3.50** | 0.71 | - | -0.34 | - | **0.96** |
| | Wordmix(En) | 79.39 | 91.86 | - | 68.71 | 96.2 | 82.62 | - | 87.57 | 91.10 | - | 79.21 | 54.97 | - | **84.29** | - | 81.59 |
| | | 0.92 | -0.69 | - | -12.66 | -0.50 | -5.61 | - | -1.21 | -1.83 | - | -1.68 | -2.03 | - | **0.86** | - | -2.37 |
| **IndicCOPA** | Native script | 65.56 | 71.11 | - | 71.43 | 78.79 | 63.33 | 61.11 | 69.32 | 63.77 | 72.86 | **73.75** | 85.51 | 61.19 | 67.50 | 84.06 | 70.66 |
| | Transliteration | 51.11 | 51.11 | - | 58.02 | 74.07 | 52.22 | 58.89 | 53.33 | 53.09 | 56.67 | 56.67 | 65.56 | 46.67 | 50.00 | 71.11 | 57.04 |
| | | -22.04 | -28.13 | - | -18.77 | -5.99 | -17.54 | -3.63 | -23.07 | -16.75 | -22.22 | -23.16 | -23.33 | -23.73 | -25.93 | -15.41 | -19.28 |
| | Translation(GPT-4) | 29.55 | 43.18 | - | 41.98 | 53.09 | 35.23 | 23.86 | 33.33 | 40.74 | 44.32 | 32.22 | 47.19 | 42.05 | 40.45 | 53.41 | 40.04 |
| | | -54.93 | -39.28 | - | -41.23 | -32.62 | -44.37 | -60.96 | -51.92 | -36.11 | -39.17 | -56.31 | -44.81 | -31.28 | -40.07 | -36.46 | -43.33 |
| | Translation(IndicTrans) | **74.44** | 65.56 | - | 65.43 | 60.00 | 61.78 | 63.64 | 62.22 | 64.56 | 56.18 | 62.22 | 66.29 | **74.44** | 67.90 | 64.00 | 65.33 |
| | | **13.54** | -7.80 | - | -8.40 | -23.85 | 7.03 | 4.14 | -10.24 | 1.24 | -22.89 | -15.63 | -22.48 | **21.65** | 0.59 | -23.86 | -7.54 |
| | Translation(implicit) | 71.26 | **78.89** | - | **72.5** | **88.33** | **77.78** | **73.53** | **75.56** | **77.97** | **81.01** | 71.59 | **89.04** | 70.37 | **75.00** | **88.46** | **77.95** |
| | | 8.69 | **10.94** | - | **1.50** | **12.11** | **22.82** | **20.32** | **9.00** | **22.27** | **11.19** | -2.93 | **4.13** | 15.00 | **11.11** | **5.23** | **10.32** |
| | Wordmix(En) | 63.33 | 77.78 | - | 66.25 | 82.81 | 68.89 | 67.78 | 74.44 | 64.20 | 65.38 | 63.33 | 73.86 | 64.29 | 74.44 | 83.33 | 70.72 |
| | | -3.40 | 9.38 | - | -7.25 | 5.10 | 8.78 | 10.91 | 7.39 | 0.67 | -10.27 | -14.13 | -13.62 | 5.07 | 10.28 | -0.87 | 0.08 |
| **IndicXNLI** | Native script | 60.85 | 68.69 | - | 62.88 | 73.28 | 65.21 | - | 62.85 | 62.67 | - | 60.55 | 68.51 | 63.98 | 62.12 | - | 64.69 |
| | Transliteration | 47.29 | 54.83 | - | 57.34 | 67.82 | 53.04 | - | 50.46 | 56.39 | - | 50.32 | 60.50 | 44.42 | 52.82 | - | 54.11 |
| | | -22.28 | -20.18 | - | -8.81 | -7.45 | -18.66 | - | -19.71 | -10.02 | - | -16.90 | -11.69 | -30.57 | -14.97 | - | -16.35 |
| | Translation(GPT-4) | 51.89 | 67.14 | - | 63.07 | 73.86 | 64.19 | - | 62.79 | 58.90 | - | 55.04 | 64.92 | 57.16 | 57.36 | - | 61.48 |
| | | -14.72 | -2.26 | - | 0.30 | 0.79 | -1.56 | - | -0.10 | -6.02 | - | -9.10 | -5.24 | -10.66 | -7.66 | - | -4.96 |
| | Translation(IndicTrans) | **65.60** | **73.23** | - | **73.27** | **76.40** | **71.22** | - | **69.49** | **67.52** | - | **69.75** | **73.95** | **69.88** | **70.49** | - | **70.98** |
| | | **7.81** | **6.61** | - | **16.52** | **4.26** | **9.22** | - | **10.56** | **7.74** | - | **15.19** | **7.94** | **9.22** | **13.47** | - | **9.72** |
| | Translation(implicit) | 62.89 | 71.03 | - | 66.75 | 75.03 | 67.69 | - | 65.36 | 64.08 | - | 63.44 | 70.98 | 66.23 | 63.52 | - | 67.00 |
| | | 3.35 | 3.41 | - | 6.15 | 2.39 | 3.80 | - | 3.99 | 2.25 | - | 4.77 | 3.61 | 3.52 | 2.25 | - | 3.57 |
| | Wordmix(En) | 55.89 | 63.98 | - | 57.89 | 71.75 | 56.44 | - | 58.95 | 60.56 | - | 52.08 | 63.92 | 59.29 | 55.36 | - | 59.65 |
| | | -8.15 | -6.86 | - | -7.94 | -2.09 | -13.45 | - | -6.21 | -3.37 | - | -13.99 | -6.70 | -7.33 | -10.88 | - | -7.79 |
| **Alexa Intent Classification** | Native script | - | 72.33 | - | - | 75.63 | 69.23 | - | 70.34 | - | - | - | 68.04 | 71.91 | 71.70 | - | 71.31 |
| | Transliteration | - | 58.07 | - | - | 73.81 | 53.07 | - | 45.01 | - | - | - | 26.85 | 55.45 | 68.39 | - | 54.38 |
| | | - | -19.72 | - | - | -2.41 | -23.34 | - | -36.01 | - | - | - | -60.54 | -22.89 | -4.62 | - | -23.74 |
| | Translation(GPT-4) | - | 75.21 | - | - | 78.57 | 73.02 | - | 72.47 | - | - | - | 70.24 | 73.00 | 74.69 | - | 73.89 |
| | | - | 3.98 | - | - | 3.89 | 5.47 | - | 3.03 | - | - | - | 3.23 | 1.52 | 4.17 | - | 3.62 |
| | Translation(IndicTrans) | - | **77.59** | - | - | 78.46 | **76.79** | - | **77.97** | - | - | - | **75.05** | **77.29** | **75.42** | - | **76.94** |
| | | - | **7.27** | - | - | 3.74 | **10.92** | - | **10.85** | - | - | - | **10.30** | **7.48** | **5.19** | - | **7.90** |
| | Translation(implicit) | - | 75.34 | - | - | **79.21** | 75.76 | - | 72.85 | - | - | - | 72.69 | 74.79 | 76.58 | - | 75.32 |
| | | - | 4.16 | - | - | **4.73** | 9.43 | - | 3.57 | - | - | - | 6.83 | 4.01 | 6.81 | - | 5.62 |
| | Wordmix(En) | - | 73.38 | - | - | 77.99 | 70.25 | - | 71.28 | - | - | - | 70.27 | 72.49 | 72.44 | - | 72.59 |
| | | - | 1.45 | - | - | 3.12 | 1.47 | - | 1.34 | - | - | - | 3.28 | 0.81 | 1.03 | - | 1.79 |

Table 10: Accuracy of all techniques for all classification datasets and languages. Below each accuracy value, the change in performance (in terms of %) wrt the native script is shown. The best results are in **boldface** and underlined. The last column shows the average performance across languages.

| | Bn | Gu | Hi | Mr | Ne | Pa | Ta | Te | Ur | | Avg |
|---|---|---|---|---|---|---|---|---|---|---|---|
| **Summarization** | | | | | | | | | | | |
| Native | 13.11 | **11.82** | **17.65** | 9.88 | 13.14 | 18.75 | **10.75** | **11.03** | 21.42 | | 14.17 |
| Transliteration | 10.16 | 10.41 | 16.08 | 8.45 | 10.87 | 12.09 | 6.13 | 7.42 | 17.57 | | 11.02 |
| Wordmix(En) | 13.11 | 8.99 | 16.62 | 9.67 | 11.73 | 16.36 | 9.66 | 8.38 | 19.41 | | 12.66 |
| Translation(IndicTrans) | 8.26 | 7.58 | 13.49 | 6.66 | 7.03 | 12.86 | 5.20 | 5.48 | 17.37 | | 9.33 |
| Translation(GPT-4) | 8.69 | 7.87 | 13.17 | 7.08 | 7.21 | 13.46 | 5.23 | 5.09 | 16.89 | | 9.41 |
| Translation(implicit) | 13.00 | 11.32 | 17.23 | **11.07** | **14.58** | **20.12** | 10.68 | 10.82 | **22.59** | | **14.60** |

| | As | Bn | Gu | Kn | Ml | Mr | Or | Pa | Ta | Te | |
|---|---|---|---|---|---|---|---|---|---|---|---|
| **MT** | | | | | | | | | | | |
| Native | 7.60 | 11.92 | 14.99 | 9.74 | 7.36 | 9.07 | 11.91 | 26.78 | 10.21 | 10.20 | 11.98 |
| Transliteration | 6.57 | 10.55 | 13.88 | 8.18 | 6.37 | 7.50 | 11.15 | 25.89 | 9.83 | 8.99 | 10.89 |
| Wordmix(En) | 7.02 | 11.74 | 14.96 | 10.08 | 7.04 | 8.99 | **12.32** | 26.97 | 10.18 | 9.49 | 11.88 |
| Translation(IndicTrans) | **8.44** | **13.15** | 15.30 | 10.23 | 7.15 | 9.09 | 10.26 | 21.52 | 9.91 | **10.49** | 11.55 |
| Translation(GPT-4) | 7.92 | 10.03 | **15.54** | 9.54 | 5.56 | 8.67 | 9.18 | 20.59 | 9.26 | 10.44 | 10.67 |
| Translation(implicit) | 7.74 | 11.72 | 15.02 | **10.36** | **7.80** | **9.71** | 12.12 | **27.26** | **10.42** | 9.06 | **12.12** |

| | Bn | Hi | Ta | Te | | Avg |
|---|---|---|---|---|---|---|
| **QA** | | | | | | |
| Native | 90.17 | **92.19** | 83.74 | 84.84 | | 87.74 |
| Transliteration | 85.37 | 86.11 | 78.50 | 91.97 | | 85.49 |
| Wordmix(En) | 80.42 | 77.93 | 78.27 | 77.70 | | 78.58 |
| Translation(IndicTrans) | 83.49 | 87.51 | **87.19** | 81.65 | | 84.96 |
| Translation(GPT-4) | 76.14 | 82.66 | 65.48 | 72.48 | | 74.19 |
| Translation(implicit) | **94.39** | 90.77 | 85.68 | **94.51** | | **91.34** |

| | Ar | Bn | Te | Th | Sw | Zh | Avg |
|---|---|---|---|---|---|---|---|
| **Reasoning** | | | | | | | |
| Native | 92.98 | 42.99 | 31.08 | 46.01 | 44.02 | 32.90 | 48.33 |
| Transliteration | 73.42 | 23.27 | 14.05 | 6.44 | - | 25.16 | 28.47 |
| Wordmix(En) | 92.31 | 37.50 | 30.80 | 34.98 | 40.00 | **35.88** | 45.25 |
| Translation(IndicTrans) | 88.33 | 43.39 | **47.33** | 31.92 | 41.92 | 31.98 | 47.48 |
| Translation(GPT-4) | 91.25 | **49.38** | 38.43 | **47.08** | **50.82** | 29.56 | **51.09** |
| Translation(implicit) | **93.67** | 42.99 | 32.74 | 40.34 | 46.15 | 32.41 | 48.05 |

Table 11: Performance(BLEU for MT/ROUGE for Summarization/EM for QA and Mathematical Reasoning) of all techniques for all generation datasets and languages. Below each performance value, the change in performance (in terms of %) wrt the native script is shown. The best results are in **boldface** and underlined. The last column shows the average performance across languages.

| Dataset | Techniques | As | Bn | Bd | Gu | Hi | Kn | Ko | Ml | Mr | Ne | Od | Pn | Ta | Te | Ur | Avg |
|---|---|---|---|---|---|---|---|---|---|---|---|---|---|---|---|---|---|
| **IndicSentiment** | Native script | 169.75 | 152.87 | 138.45 | 204.75 | 131.15 | 251.00 | - | 241.16 | 136.43 | - | 318.29 | 208.42 | 203.76 | 241.91 | 125.50 | 194.11 |
| | Transliteration | 51.29 | 48.40 | 55.82 | 50.61 | 50.84 | 57.46 | - | 60.01 | 52.72 | - | 53.25 | 54.52 | 52.69 | 49.27 | 58.56 | 53.50 |
| | | -69.78 | -68.34 | -59.68 | -75.28 | -61.24 | -77.11 | - | -75.12 | -61.36 | - | -83.27 | -73.84 | -74.14 | -79.63 | -53.34 | -72.44 |
| | Translation(GPT-4) | 226.25 | 208.51 | 199.09 | 260.73 | 187.23 | 305.94 | - | 296.78 | 194.23 | - | 373.73 | 263.94 | 259.40 | 297.93 | 181.56 | 250.41 |
| | | 33.28 | 36.40 | 43.80 | 27.34 | 42.76 | 21.89 | - | 23.06 | 42.37 | - | 17.42 | 26.64 | 27.31 | 23.16 | 44.67 | 29.00 |
| | Translation(IndicTrans) | **27.68** | **27.25** | **27.95** | **27.19** | **27.11** | **27.10** | - | **27.18** | **27.98** | - | **26.67** | **27.18** | **26.81** | **27.65** | **26.55** | **27.25** |
| | | -83.69 | -82.17 | -79.81 | -86.72 | -79.33 | -89.20 | - | -88.73 | -79.49 | - | -91.62 | -86.96 | -86.84 | -88.57 | -78.84 | -85.96 |
| | Wordmix(En) | 140.10 | 135.95 | 126.83 | 131.85 | 120.43 | 179.47 | - | 199.11 | 124.12 | - | 203.23 | 157.81 | 184.71 | 176.73 | 110.28 | 153.12 |
| | | -17.47 | -11.07 | -8.39 | -35.60 | -8.17 | -28.5 | - | -17.44 | -9.02 | - | -36.15 | -24.28 | -9.35 | -26.94 | -12.13 | -21.12 |
| **IndicXParaphrase** | Native script | 263.87 | 218.26 | - | 300.39 | 189.15 | 345.18 | - | 328.80 | 201.89 | - | 482.65 | 298.38 | - | 322.54 | - | 295.11 |
| | Transliteration | 81.27 | 68.20 | - | 73.43 | 78.25 | 80.37 | - | 83.80 | 81.92 | - | 78.83 | 74.17 | - | 69.29 | - | 76.95 |
| | | -69.20 | -68.75 | - | -75.56 | -58.63 | -76.72 | - | -74.51 | -59.42 | - | -83.67 | -75.14 | - | -78.52 | - | -73.92 |
| | Translation(GPT-4) | 361.11 | 311.40 | - | 395.81 | 287.61 | 440.52 | - | 421.46 | 298.71 | - | 576.73 | 391.70 | - | 413.86 | - | 389.89 |
| | | 36.85 | 42.67 | - | 31.77 | 52.05 | 27.62 | - | 28.18 | 47.96 | - | 19.49 | 31.28 | - | 28.31 | - | 32.12 |
| | Translation(IndicTrans) | **46.84** | **45.21** | - | **44.82** | **46.38** | **45.87** | - | **44.72** | **45.65** | - | **45.48** | **45.42** | - | **43.94** | - | **45.43** |
| | | -82.25 | -79.29 | - | -85.08 | -75.48 | -86.71 | - | -86.40 | -77.39 | - | -90.58 | -84.78 | - | -86.38 | - | -84.61 |
| | Wordmix(En) | 222.84 | 193.80 | - | 194.98 | 167.77 | 245.38 | - | 275.81 | 172.48 | - | 327.88 | 216.30 | - | 207.18 | - | 222.44 |
| | | -15.55 | -11.21 | - | -35.09 | -11.30 | -28.91 | - | -16.12 | -14.57 | - | -32.07 | -27.51 | - | -35.77 | - | -24.62 |
| **IndicCOPA** | Native script | 131.50 | 123.83 | - | 140.07 | 93.01 | 176.83 | 98.20 | 166.03 | 90.37 | 92.72 | 229.17 | 149.84 | 162.91 | 173.98 | 82.74 | 136.51 |
| | Transliteration | 42.14 | 41.24 | - | 37.20 | 39.58 | 44.03 | 41.21 | 44.80 | 38.48 | 39.58 | 38.68 | 40.40 | 40.59 | 38.60 | 38.89 | 40.39 |
| | | -67.95 | -66.70 | - | -73.44 | -57.45 | -75.10 | -58.03 | -73.02 | -57.42 | -57.31 | -83.12 | -73.04 | -75.08 | -77.81 | -53.00 | -70.41 |
| | Translation(GPT-4) | 180.16 | 171.23 | - | 185.21 | 137.63 | 221.43 | 145.56 | 209.85 | 135.09 | 136.56 | 272.93 | 195.44 | 209.51 | 218.68 | 126.78 | 181.85 |
| | | 37.00 | 38.28 | - | 32.23 | 47.97 | 25.22 | 48.23 | 26.39 | 49.49 | 47.28 | 19.09 | 30.43 | 28.6 | 25.69 | 53.23 | 33.21 |
| | Translation(IndicTrans) | **21.82** | **21.92** | - | **21.70** | **21.77** | **21.76** | **22.84** | **21.59** | **21.49** | **21.91** | **21.40** | **22.28** | **21.03** | **21.81** | **21.64** | **21.78** |
| | | -83.41 | -82.30 | - | -84.51 | -76.59 | -87.69 | -76.74 | -87.00 | -76.22 | -76.37 | -90.66 | -85.13 | -87.09 | -87.46 | -73.85 | -84.05 |
| | Wordmix(En) | 111.64 | 105.96 | - | 99.88 | 83.05 | 140.72 | 89.53 | 144.91 | 80.42 | 91.86 | 168.23 | 102.78 | 142.26 | 149.21 | 74.59 | 113.22 |
| | | -15.10 | -14.43 | - | -28.69 | -10.71 | -20.42 | -8.83 | -12.72 | -11.01 | -0.93 | -26.59 | -31.41 | -12.68 | -14.24 | -9.85 | -17.06 |
| **IndicXNLI** | Native script | 202.37 | 187.22 | - | 243.68 | 155.95 | 287.14 | - | 270.63 | 155.30 | - | 384.11 | 246.97 | 240.77 | 264.73 | - | 239.90 |
| | Transliteration | 62.43 | 59.87 | - | 61.09 | 66.20 | 67.33 | - | 67.87 | 61.97 | - | 63.64 | 63.61 | 63.55 | 57.34 | - | 63.17 |
| | | -69.15 | -68.02 | - | -74.93 | -57.55 | -76.55 | - | -74.92 | -60.10 | - | -83.43 | -74.24 | -73.61 | -78.34 | - | -73.67 |
| | Translation(GPT-4) | 270.33 | 255.62 | - | 311.00 | 223.25 | 352.72 | - | 334.49 | 220.40 | - | 449.53 | 313.43 | 306.19 | 329.47 | - | 306.04 |
| | | 33.58 | 36.53 | - | 27.63 | 43.15 | 22.84 | - | 23.6 | 41.92 | - | 17.03 | 26.91 | 27.17 | 24.46 | - | 27.57 |
| | Translation(IndicTrans) | **32.10** | **32.38** | - | **32.44** | **33.59** | **32.14** | - | **31.02** | **31.77** | - | **31.97** | **32.56** | **32.53** | **32.23** | - | **32.25** |
| | | -84.14 | -82.70 | - | -86.69 | -78.46 | -88.81 | - | -88.54 | -79.54 | - | -91.68 | -86.82 | -86.49 | -87.83 | - | -86.56 |
| | Wordmix(En) | 179.39 | 170.82 | - | 172.71 | 142.24 | 194.88 | - | 227.87 | 138.80 | - | 266.72 | 181.60 | 210.34 | 175.86 | - | 187.38 |
| | | -11.36 | -8.76 | - | -29.12 | -8.79 | -32.13 | - | -15.80 | -10.62 | - | -30.56 | -26.47 | -12.64 | -33.57 | - | -21.89 |
| **Alexa Intent Classification** | Native script | - | 41.64 | - | - | 34.59 | 67.53 | - | 70.01 | - | - | - | 57.60 | 66.10 | - | 32.00 | 52.78 |
| | Transliteration | | 12.52 | | | 14.50 | 14.52 | | 15.84 | | | | 13.97 | 12.96 | | 15.10 | 14.20 |
| | | | -69.93 | | | -58.08 | -78.50 | | -77.37 | | | | -75.75 | -80.39 | | -52.81 | -73.10 |
| | Translation(GPT-4) | | 58.76 | | | 51.93 | 84.41 | | 87.31 | | | | 74.74 | 83.56 | | 49.38 | 70.02 |
| | | | 44.11 | | | 50.13 | 25.00 | | 24.71 | | | | 29.76 | 26.41 | | 54.31 | 32.66 |
| | Translation(IndicTrans) | | **7.56** | | | **7.68** | **7.41** | | **7.55** | | | | **7.37** | **7.61** | | **7.53** | **7.53** |
| | | | -81.84 | | | -77.80 | -89.03 | | -89.22 | | | | -87.20 | -88.49 | | -76.47 | -85.73 |
| | Wordmix(En) | | 35.59 | | | 31.79 | 47.63 | | 62.98 | | | | 49.17 | 46.55 | | 28.56 | 43.18 |
| | | | -14.53 | | | -8.09 | -29.47 | | -10.04 | | | | -14.64 | -29.58 | | -10.75 | -18.19 |

Table 12: Average tokens generated per instance of all techniques for all classification datasets and languages. Below each average token value, the change in token generation (in terms of %) wrt the native script is shown. The best results are in **boldface** and underlined. The last column shows the average cost across languages.

**Summarization**

| | Bn | Gu | Hi | Mr | Ne | Pa | Ta | Te | Ur | Avg |
|---|---|---|---|---|---|---|---|---|---|---|
| **Native** | 4187.78 | 4677.24 | 2500.08 | 3769.24 | 2682.88 | 4262.24 | 3854.80 | 4836.57 | 1948.24 | 3635.45 |
| **Transliteration** | 1347.94 | 1244.39 | 1056.28 | 1518.29 | 1093.74 | 1062.06 | 1142.00 | 1279.42 | 937.04 | 1186.80 |
| **Wordmix(En)** | 3709.58 | 3497.39 | 2226.51 | 3342.76 | 2428.56 | 3037.89 | 3568.86 | 4177.00 | 1737.97 | 3080.72 |
| **Translation(IndicTrans)** | **_718.69_** | **_704.24_** | **_522.46_** | **_813.13_** | **_548.78_** | **_595.87_** | **_592.12_** | **_776.61_** | **_473.06_** | **_638.33_** |
| **Translation(GPT-4)** | 4696.44 | 5180.26 | 2991.28 | 4269.80 | 3190.36 | 4769.94 | 4355.26 | 5337.75 | 2433.48 | 4136.06 |

**MT**

| | As | Bn | Gu | Kn | Ml | Mr | Or | Pa | Ta | Te | |
|---|---|---|---|---|---|---|---|---|---|---|---|
| **Native** | 70.95 | 65.47 | 50.31 | 45.30 | 57.52 | 72.00 | 53.29 | 78.34 | 69.14 | 53.66 | 61.60 |
| **Transliteration** | 31.16 | 29.16 | 20.57 | 18.86 | 24.25 | 30.93 | 22.96 | 33.87 | 30.13 | 22.96 | 26.49 |
| **Wordmix(En)** | 70.95 | 65.47 | 50.31 | 44.10 | 56.79 | 64.06 | 53.29 | 78.33 | 69.14 | 54.02 | 60.65 |
| **Translation(IndicTrans)** | **_15.18_** | **_15.11_** | **_11.80_** | **_10.23_** | **_13.37_** | **_16.52_** | **_12.41_** | **_18.29_** | **_15.33_** | **_12.92_** | **_14.12_** |
| **Translation(GPT-4)** | 101.65 | 94.99 | 74.33 | 67.54 | 85.44 | 104.98 | 79.63 | 115.82 | 100.62 | 79.88 | 90.49 |

**QA**

| | Bn | Hi | Ta | Te | Avg |
|---|---|---|---|---|---|
| **Native** | 834.50 | 898.04 | 4269.58 | 843.30 | 1711.36 |
| **Transliteration** | 266.27 | 388.81 | 1240.61 | 202.65 | 524.59 |
| **Wordmix(En)** | 706.96 | 804.61 | 2848.23 | 536.98 | 1224.20 |
| **Translation(IndicTrans)** | **_176.85_** | **_226.42_** | **_662.53_** | **_153.54_** | **_304.84_** |
| **Translation(GPT-4)** | 1161.32 | 1305.72 | 4765.46 | 1124.10 | 2089.15 |

**Reasoning**

| | Ar | Bn | Te | Th | Sw | Zh | Avg |
|---|---|---|---|---|---|---|---|
| **Native** | 50.82 | 282.13 | 454.11 | 199.31 | 102.09 | 77.96 | 194.40 |
| **Transliteration** | 32.59 | 95.20 | 96.21 | 89.78 | - | 105.89 | 83.93 |
| **Wordmix(En)** | 40.42 | 228.37 | 247.47 | 125.36 | 97.92 | 76.15 | 135.95 |
| **Translation(IndicTrans)** | **_22.25_** | **_57.67_** | **_58.66_** | **_49.32_** | **_58.49_** | **_73.39_** | **_53.30_** |
| **Translation(GPT-4)** | 98.64 | 400.67 | 569.21 | 318.75 | 220.29 | 222.50 | 305.01 |

Table 13: Average tokens generated per instance of all techniques for all generation datasets and languages. Below each average token value, the change in token generation (in terms of %) wrt the native script is shown. The best results are in **boldface** and underlined. The last column shows the average cost across languages.